\documentclass{article}

\usepackage{arxiv}

\usepackage[utf8]{inputenc} 
\usepackage[T1]{fontenc}    
\usepackage{hyperref}       
\usepackage{url}            
\usepackage{booktabs}       
\usepackage{amsfonts}       
\usepackage{nicefrac}       
\usepackage{microtype}      
\usepackage{lipsum}
\usepackage{graphicx}
\usepackage{hyperref}
\usepackage{epstopdf, epsfig}
\usepackage{bm}
\usepackage{amsthm}
\usepackage{amsmath,amssymb}
\usepackage[utf8]{inputenc}
\usepackage[T1]{fontenc}
\usepackage{mathptmx}
\usepackage[numbers]{natbib}
\usepackage{color}
\usepackage{subfigure}
\usepackage[ruled,linesnumbered]{algorithm2e}
\SetKwComment{Comment}{$\triangleright$\ }{}

\usepackage{newfloat,algcompatible}
\usepackage[size=small]{caption}

\title{Energy networks for state estimation with random sensors using sparse labels}

\author{
  Yash Kumar \\
  Department of Mechanical Engineering\\
  Delhi Technological University\\
  Shahbad Daulatpur, Main Bawana Road, Delhi-110042, India \\
  \texttt{yashk8481@gmail.com} \\
   \And
 Souvik Chakraborty \\
  Department of Applied Mechanics\\
  Indian Institute of Technology Delhi\\
  Hauz Khas - 110042, New Delhi, India \\
  \texttt{souvik@am.iitd.ac.in} \\
}

\begin{document}
\maketitle

\begin{abstract}
State estimation is required whenever we deal with high-dimensional dynamical systems, as the complete measurement is often unavailable. It is key to gaining insight, performing control or optimizing design tasks. Most deep learning-based approaches require high-resolution labels and work with fixed sensor locations, thus being restrictive in their scope. Also, doing Proper orthogonal decomposition (POD) on sparse data is nontrivial. To tackle these problems, we propose a technique with an implicit optimization layer and a physics-based loss function that can learn from sparse labels. It works by minimizing the energy of the neural network prediction, enabling it to work with a varying number of sensors at different locations. Based on this technique we present two models for discrete and continuous prediction in space. We demonstrate the performance using two high-dimensional fluid problems of Burgers’ equation and Flow Past Cylinder for discrete model and using Allen–Cahn equation and Convection-diffusion equations for continuous model. We show the models are also robust to noise in measurements.
\end{abstract}

\keywords{State estimation \and Differentiable implicit layers \and Dynamical systems}

\section{\label{sec:level1}Introduction}
    State estimation is the ability to recover flow based on a few measurements. It is an inverse problem and arises in many engineering applications such as remote sensing, medical imaging, ocean dynamics, reservoir modeling, and blood flow modeling. Uses of fluid estimation include flow control \cite{Cordier2011, Semaan}, cardiac blood flow modeling \cite{pmid22539149, pmid18092738, KISSAS2}, ship wake identification \cite{GRAZIANO201672}, climate prediction \cite{kalnay_2002}, optimizing machine design for low-drag vehicles, efficient turbo-machines, etc. Few challenges faced in the processes are limited sensors, sparse label data, moving sensors, ill-posed problems, noisy measurements, etc. This work focuses on learning from moving sparse label data and sensor measurements with a deep learning-based model using an implicit optimization layer for network training. Sparse fluid is encountered in various situations. One reason is that storing high-resolution data generated during direct numerical simulation is challenging due to limited storage space. It makes analysis, sharing, and visualization difficult. Another important reason is that real data is hard to measure on a full scale, like cardiovascular blood flow data obtained from flow magnetic resonance imaging (MRI) \cite{Ong, pmid27885707, pmid30423501}.
    
    For high dimensional state estimation problems like fluids, popular approaches include library-based approaches observer dynamical system stochastic approaches. Library-based methods use offline data, and the library consists of generic modes such as Fourier, wavelet, discrete cosine transform basis, or data specific Proper orthogonal decomposition (POD) or Dynamic mode decomposition (DMD) modes, or training data. Library-based approaches using sparse representation assume state can be expressed as the combination of library elements. 
    In an observer dynamical system, we assume the system's dynamics to produce a full state and update it based on new measurements to reduce estimation error forming a closed feedback loop. The estimate is maintained by Kalman filtering \cite{bishop2006pattern, reif1999stochastic, wan2001unscented}. \citet{Jonathan} applied dynamic mode decomposition \cite{2008AP, rowley_mez} as a reduced-order model to Kalman smoother estimate to identify coherent structures. \citet{BUFFONI20082626} used a nonlinear observer-based on Galerkin projection of Navier-Stokes equation to estimate POD coefficients. 
    Stochastic estimation was proposed by \citet{Adrian} for a turbulence study where the conditional mean was approximated using a power series. Extension to this method \cite{Guezennec, Ewing, Naguib} can be found in the literature. \citet{Bonnet1994} extended stochastic approach to estimate POD coefficients. A linear mapping between sensors and coefficients was assumed. These approaches allow more flexibility in sensor placements and have been applied for flow control over airfoil \cite{Aero} and analyzing isotropic turbulence \cite{tur, Tung}.
    
    We consider a problem with sparse label data whose position may vary with time, for which POD can not be performed. Thus it restricts us from using traditional POD-based approaches. Other deep learning-based approaches like \cite{kumar2021state, erichson2019shallow, nair_goza_2020} give accurate predictions with fewer sensors but require high-resolution labels for training. \citet{gaoHan} uses physics-based loss for super-resolution using spare data and thus assumes a fixed number of sensors and their positions. In this work, we propose an Energy network for state estimation with random sensors (ENSERS), a technique to learn a model from spare training labels capable of predicting full states given a varied number of sensors at random locations. We present two models trained using this technique. The first one produces discrete high-dimensional predictions in space. Second, produce continuous prediction utilizing the information of coordinates. We demonstrate the results corresponding to four high complexity problems: 2-dimensional (2D) coupled Burgers’ equation, transient flow, Allen–Cahn equation, and Convection-diffusion equation.

    The remainder of the paper is organized as follows. In Section \ref{sec:ps}, details on the problem statement is provided. Details on the proposed approach are provided in Section \ref{sec:pa}. Section \ref{sec:dsm} and \ref{sec:csm} give details on discrete and continuous formulation with two numerical examples in each to illustrate the performance of the proposed approach. Finally, Section \ref{sec:conc} provides the concluding remarks.

\section{Problem statement}\label{sec:ps}

    Consider a dynamical system obtained by partial discretization of the $d$-dimensional governing differential equations:
    \begin{equation}
        J_t(x, t) = F(x, J(x, t)), \quad J^n = J(x, t_n), \quad x \in \Omega
        \label{eq:system_eq}
    \end{equation}
    The simulation time domain is discretized by $L$ steps and the space domain is discretized by $\omega$ segments resulting in
    \begin{equation}
        \label{eq:simData}
        Z = \{ J^l_m \in \mathbb R^{\omega} \ | \ l=0,...,L-1, \ m=0,...,M-1 \}
    \end{equation}
    where $Z \in \mathbb R^{L \times M \times \omega}$, $l$ is time step index, $M=$ number of system’s state variables. e.g. $m=0$ represents x-velocity and $m=1$ represents y-velocity in $\S$ \ref{sec:burgers} of 2D coupled Burgers’ equation. We consider sensor location and data location, represented by integers in discrete domain respectively as  
    \begin{equation}
        \label{eq:senLocEachT}
        S = \{ \lambda^l_m \in \mathbb Z^{p} \cap [0, \omega-1]^{p} \ | \ l=0,...,L-1, \ m=0,...,M-1 \}
    \end{equation}
    \begin{equation}
        \label{eq:dataLocEachT}
        T = \{ \pi^l_m \in \mathbb Z^{h} \cap [0, \omega-1]^{h} \ | \ l=0,...,L-1, \ m=0,...,M-1 \}
    \end{equation}
    where $p=$ number of sensors, $h=$ number of data nodes, $S \in \mathbb Z^{L \times M \times p} \cap [0, \omega-1]^{L \times M \times p}$, $T \in \mathbb Z^{L \times M \times h} \cap [0, \omega-1]^{L \times M \times h}$. The corresponding sensor values and data values are 
    
    \begin{equation}
        X = \{ \sigma^l_m \in \mathbb R^{p} \ | \ l=0,...,L-1, \ m=0,...,M-1 \}
    \end{equation}
    \begin{equation}
        \Phi = \{ \psi^l_m \in \mathbb R^{h} \ | \ l=0,...,L-1, \ m=0,...,M-1 \}
    \end{equation}
    
    \begin{equation}
        \sigma^l_m = \Lambda^{l}_m J^l_m
    \end{equation}
    \begin{equation}
        \psi^l_m = \Pi^{l}_m J^l_m
    \end{equation}
    where $X \in \mathbb R^{L \times M \times p}$, $\Phi \in \mathbb R^{L \times M \times h}$. $\Lambda^{l}_m \in \mathbb R^{p \times \omega}$ and $\Pi^{l}_m  \in \mathbb R^{h \times \omega}$ are measurement matrices composed of one-hot row vectors. $\Lambda^{l}_m$ and $\Pi^{l}_m$ are defined as
    
    \begin{equation}
        \Lambda^{l}_{mij} = \left\{ 
                              \begin{array}{ c l }
                                1 & \quad \textrm{if } \lambda^l_{mi} = j \\
                                0                 & \quad \textrm{otherwise}
                              \end{array}
                            \right.
    \end{equation}
    \begin{equation}
        \Pi^{l}_{mij} = \left\{ 
                              \begin{array}{ c l }
                                1 & \quad \textrm{if } \pi^l_{mi} = j \\
                                0                 & \quad \textrm{otherwise}
                              \end{array}
                            \right.
    \end{equation}
    Note that integer senor locations $\lambda^l_m$ are selected randomly and are kept fixed during training. Similarly, a different set of sensor locations $S$ is selected for testing the network. In this work, we aim to use sensor data from a set of $\gamma$ system states and produce $\gamma$ high dimensional states. Thus we divided the sensor values and data into chunks, each with $\gamma$ states.

    \begin{equation}
        \chi^k = \{ \sigma^{k*z+i}_{m} \in \mathbb R^{p} \ | \ i=0,...,\gamma-1, \ m=0,...,M-1 \}
    \end{equation}
    
    \begin{equation}
        \phi^k = \{ \psi^{k*z+i}_{m}  \in \mathbb R^{h} \ | \ i=0,...,\gamma-1, \ m=0,...,M-1 \}
    \end{equation}
    where $\chi^k \in \mathbb R^{\gamma \times M \times p}$ are sensor values for $\gamma$ time steps, $\phi^k \in \mathbb R^{\gamma \times M \times h}$ are data values for $\gamma$ time steps, $k=0,...,N-1$, $N$ is number of train samples, $z$ is time steps between first state $\sigma^{k*z}_{m}$ of each tensor $\chi^k$.

\section{Discrete space models}\label{sec:dsm}

    \subsection{Proposed approach}\label{sec:pa}
        
        \begin{figure}[!htb]
            \centering
            \includegraphics[width=1\linewidth]{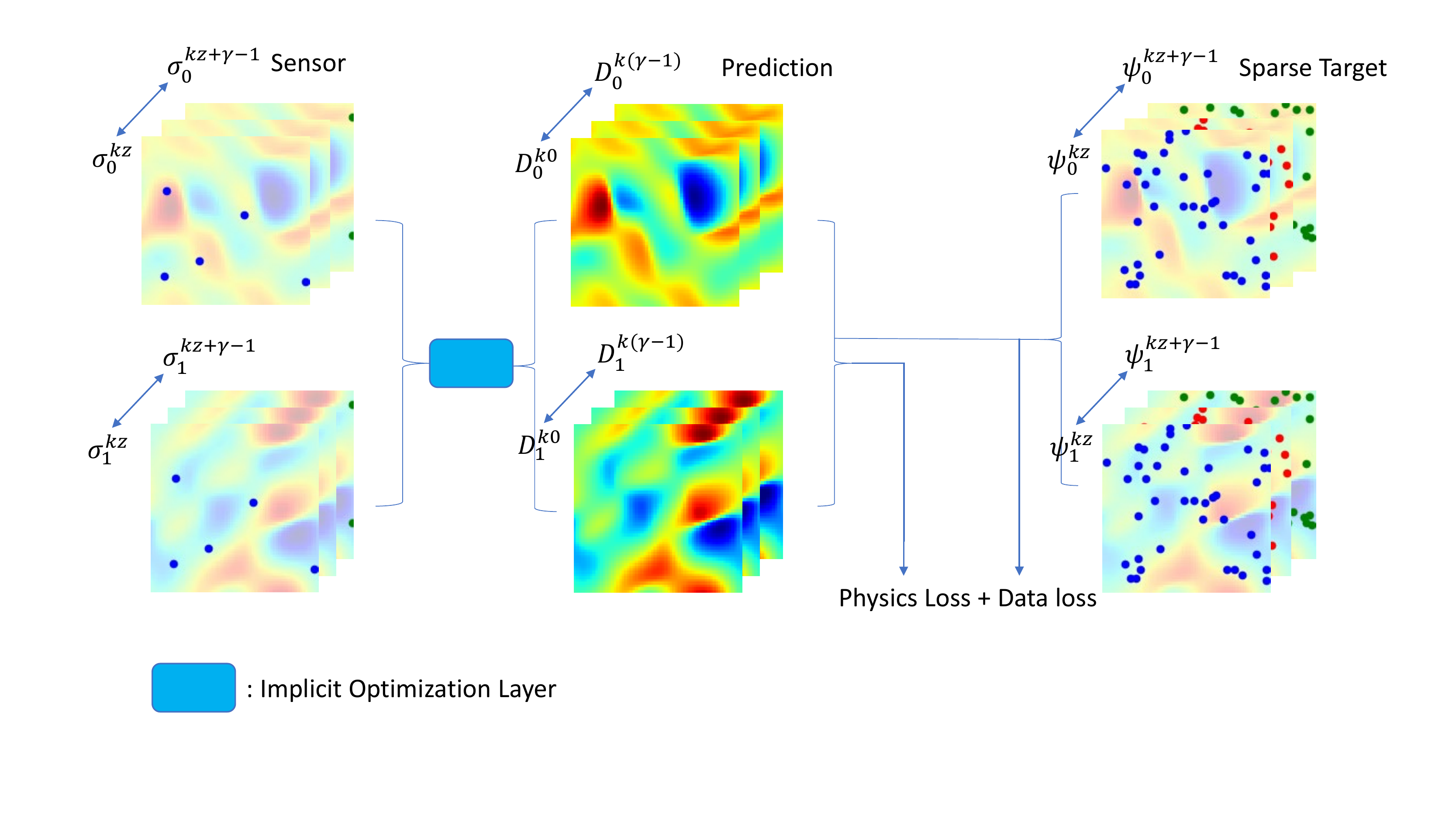}
            \caption{Network architecture of proposed Ensers model.}
            \label{fig:architecture}
        \end{figure}
        
        In this section, we propose a novel deep learning-based framework for state estimation. Prediction of $\gamma$ high dimensional states is done via feed-forward neural network (FNN) $\Gamma$ using optimized reduced state vector $\xi^k \in \mathbb R^{\varsigma}$:
    
        \begin{equation}
            \label{eq:decoder}
            \Gamma(\xi^k) = D^k = \{ D^{ki}_m \in \mathbb R^{\omega} \ | \ i=0,...,\gamma-1, \ m=0,...,M-1 \}
        \end{equation}
        where $D^k \in \mathbb R^{ \gamma \times M \times \omega} $ is a third-order tensor of predicted states. The output vector of FNN has a dimension of $ \gamma * M * \omega$ which is reshaped into a tensor of shape $\gamma \times M \times \omega$. The reduced state is obtained by solving the following minimization problem using sensor data $\chi^k$ and predicted states by the neural network.
        \begin{subequations}
            \label{eq:optimize}
            
            \begin{equation}
                \xi^k = \arg \min_{\tilde{ \xi^k }} \| vec(\chi^k - \rho( \tilde{ \xi^k } )) \|_2^2
            \end{equation}
            
            \begin{equation}
                \label{eq:rho_xi_tilde}
                \rho(\tilde{ \xi^k }) = \{ \tilde{Q}^{ki}_m \in \mathbb R^{p} \ | \ i=0,...,\gamma-1, \ m=0,...,M-1 \}
            \end{equation}
            
            \begin{equation}
                \label{eq:ValSenser}
                \tilde{Q}^{ki}_m = \Lambda^{kz+i}_m \tilde{D}^{ki}_m
            \end{equation}
            
            \begin{equation}
                \label{eq:NN_xi_tilde}
                \Gamma(\tilde{ \xi^k }) = \{ \tilde{D}^{ki}_m \in \mathbb R^{\omega} \ | \ i=0,...,\gamma-1, \ m=0,...,M-1 \}
            \end{equation}
        \end{subequations}
        where $\tilde{Q}^{ki}_m \in \mathbb R^{p}$ are values of predicted states at sensor locations $\lambda^{kz+i}_m$, $\Lambda^{kz+i}_m \in \mathbb R^{p \times \omega}$ is measurement matrix. Note that in practice $\xi^k$ is obtained by a few steps of gradient descent instead of global minimization. This inner optimization loop is implemented using the library `higher' \cite{grefen} in PyTorch. Also, dimension $\varsigma$ of the reduced state vector is quite small (e.g. 8 in first experiment $\S$ \ref{sec:burgers}) therefore the time required for gradient descent steps is negligible. The network $\Gamma$ is trained by minimizing data loss and physics-based loss $P(\chi^k)$ across training samples $N$.
        
        \begin{subequations}
            \begin{equation}
                \theta^* = \arg \min_{ \theta } \sum_{k=0}^{N-1} \| vec(\phi^k - \Upsilon( \chi^k )) \|_2^2 + P(\chi^k)
            \end{equation}
            
            \begin{equation}
                \label{eq:train_b}
                \Upsilon( \chi^k ) = \{ Y^{ki}_m \in \mathbb R^{h} \ | \ i=0,...,\gamma-1, \ m=0,...,M-1 \}
            \end{equation}
            
            \begin{equation}
                \label{eq:ValData}
                Y^{ki}_m = \Pi^{kz+i}_m D^{ki}_m
            \end{equation}
        \end{subequations}
        where $D^{ki}_m \in \mathbb R^{\omega}$ are predicted states by network $\Gamma$ from Eq. (\ref{eq:decoder}), $Y^{ki}_m \in \mathbb R^{h}$ are values of predicted states at data locations $\pi^{kz+i}_m$, $\Upsilon( \chi^k )$ are predicted states composed of $\gamma$ time steps. Fig. \ref{fig:architecture} shows the network architecture during training. Equations (\ref{eq:decoder}) and (\ref{eq:optimize}) together forms the implicit optimization layer shown in the Fig. \ref{fig:architecture}. Training and testing procedure is shown in Algorithm \ref{alg:train} and \ref{alg:test} respectively. Note that for demonstration purpose, in all Algorithms a batch size of $1$ is considered but in practice batch size is selected based on problem as mentioned in each experiment section. Also we use Huber loss function during testing as it is more robust to noise than mean squared error (MSE) loss function.
        
        Methods based on implicit optimization layers come under the category of Optimization-based Modeling architectures \cite{amos2019} and are well-studied for generic classification, and structured prediction tasks \cite{Multipred, stoyanov11a, brakel13a, Lecun}. The most common way of training such models is through Unrolled Differentiation \cite{Utama, belanger2017endtoend, Metz, stoyanov11a}. It is done by introducing an optimization procedure such as gradient descent into the inference procedure. Other ways of training include Implicit argmin differentiation using implicit function theorem but need argmin operations to be convex. This method can be found in works of \citep{Johnson, JordanSquire}. In this work, we use Unrolled Differentiation technique for training.

        \begin{algorithm}[hbt!]
        
            \caption{Training ENSERS}
            \label{alg:train}
            \begin{algorithmic}[1]
        
            \STATE{\textbf{Inputs:} $Z, S, T$.     \COMMENT{Eq. \eqref{eq:simData}, Eq. \eqref{eq:senLocEachT}, Eq. \eqref{eq:dataLocEachT}}} 
            
            \STATE{\textbf{Set Hyper-parameters:} $\eta_o$: outer learning rate, $\eta_{i0}$: inner learning rate at epoch=0, $\hat{\eta_{i}}$: inner learning rate rate, $\beta$: batch size, $I_o$: outer iterations, $I_i$: inner iterations, $\hat{\zeta}$: physics penalty rate, $\zeta_0$: physics penalty at epoch=0, $N$: number of train samples, $z$, $\gamma$.} 
            
            \STATE{\textbf{Calculate data-set:} $\chi$, $\phi$, $\Lambda$, $\Pi$}
            
            \STATE{\textbf{Initialize:} Neural network model: $ \Gamma(\cdot; \theta)$}
        
            \FOR{ $\iota_o = 0$ \textbf{to} $I_o-1$}        \COMMENT{Outer optimization Loop}
            
                \STATE{$\eta_{i} = \eta_{i0} + \iota_o \hat{\eta_{i}}$} \COMMENT{Schedule inner learning rate}
                \STATE{$\zeta = \zeta_0 + \iota_o \hat{\zeta}$}         \COMMENT{Schedule physics penalty}
                \FOR{$k = 0$ \textbf{to} $N-1$}
                
                    \STATE{\textbf{Initialize:} $\tilde{ \xi^k }$}
                    \FOR{$\iota_i = 0$ \textbf{to} $I_i-1$}        \COMMENT{Inner optimization Loop}
                        
                        \STATE{$ \tilde{D^k} = \Gamma( \tilde{\xi^k} ) $}   \COMMENT{Eq. \eqref{eq:NN_xi_tilde}}
                        \STATE{$ \tilde{Q^k} \leftarrow$ DrawValuesAtSensorLocations($\tilde{D^k}$) }   \COMMENT{Eq. \eqref{eq:ValSenser}}
                        \STATE{$\mathfrak L = MSE(  \chi^k, \tilde{Q^k} )$}
                        \STATE{$\frac{\partial \mathfrak L}{\partial \tilde{ \xi^k }} \leftarrow$ Backprop($\mathfrak L$)}
                        \STATE{$\tilde{ \xi^k } = \tilde{ \xi^k } -  \eta_i \frac{\partial \mathfrak L}{\partial \tilde{ \xi^k }} $}
                    \ENDFOR
                    \STATE{$\xi^k = \tilde{ \xi^k }$}
                    \STATE{$D^k = \Gamma(\xi^k)$   \COMMENT{Eq. \eqref{eq:decoder}}}
                    \STATE{$Y^k \leftarrow$ DrawValuesAtDataLocations($D^k$)    \COMMENT{Eq. \eqref{eq:ValData}}}
                        
                    \STATE{$\mathcal L = MSE(\phi^k, Y^k) + \zeta P(D^k)     $\COMMENT{Calculate loss}}
                    \STATE{$\frac{\partial \mathcal L}{\partial \theta} \leftarrow$ Backprop($\mathcal L$)}
                    \STATE{$ \theta \leftarrow \theta - \eta_o \frac{\partial \mathcal L}{\partial \theta}$        \COMMENT{Update weights}}
                \ENDFOR
            \ENDFOR 
        
            \STATE{\textbf{Output:} Trained network $ \Gamma(\cdot; \theta^*)$.}
        
            \end{algorithmic}
        \end{algorithm}

        \begin{algorithm}[hbt!]
            \caption{Testing ENSERS}
            \label{alg:test}
            \begin{algorithmic}[1]
        
            \STATE{\textbf{Inputs:} Trained network $ \Gamma(\cdot; \theta^*)$, $S$.  \quad    \COMMENT{Eq. \eqref{eq:senLocEachT}} }
            
            \STATE{\textbf{Set Hyper-parameters:} $\eta_i$: inner learning rate, $I_i$: inner iterations, $\hat{N}$: number of test samples, $z$, $\gamma$.}
            
            \STATE{\textbf{Calculate data-set:} $\chi$, $\Lambda$}
            
            \FOR{$k = 0$ \textbf{to} $\hat{N}-1$}
            
                \STATE{\textbf{Initialize:} $\tilde{ \xi^k }$}
                \FOR{$\iota_i = 0$ \textbf{to} $I_i-1$}        \COMMENT{Inner optimization Loop}
                    
                    \STATE{$ \tilde{D^k} = \Gamma( \tilde{\xi^k} ) $}   \COMMENT{Eq. \eqref{eq:NN_xi_tilde}}
                    \STATE{$ \tilde{Q^k} \leftarrow$ DrawValuesAtSensorLocations($\tilde{D^k}$) }   \COMMENT{Eq. \eqref{eq:ValSenser}}
                    \STATE{$\mathfrak L = MSE(  \chi^k, \tilde{Q^k}  )$}
                    \STATE{$\frac{\partial \mathfrak L}{\partial \tilde{ \xi^k }} \leftarrow$ Backprop($\mathfrak L$)}
                    \STATE{$\tilde{ \xi^k } = \tilde{ \xi^k } -  \eta_i \frac{\partial \mathfrak L}{\partial \tilde{ \xi^k }} $}
                \ENDFOR
                
                \STATE{$\xi^k = \tilde{ \xi^k }$}
                \STATE{$D^k = \Gamma(\xi^k)$     \COMMENT{Eq. \eqref{eq:decoder}}}
            \ENDFOR
             
            \STATE{\textbf{Output:} Predicted states $\{ D^{ki}_m \ | \ k=0,...,N-1, \ i=0,...,\gamma-1, \ m=0,...,M-1 \}$.}
        
            \end{algorithmic}
        \end{algorithm}

        \subsubsection{Physics-based loss function}
            The physics-based loss function is used in the approach because of spare training labels. Training any network with just spare labels will produce garbage values on nodes without labels. Physics-based loss functions have been used to train neural networks for solving PDEs. A popular class of methods is PINNs \cite{RAISSI_p}. The basic idea here is to place a neural network prior to the state variable and then estimate the neural network parameters by using a physics-informed loss function. Several improvements to the originally proposed PINN can also be found in the literature. For example, \citet{ZHU201956} developed convolutional PINN for time-independent systems. \citet{GENEVA} used physics constrained auto-regressive model for surrogate modeling of dynamical systems. We use Runge-Kutta methods with $q$ stages for defining loss between $\gamma$ state predictions. Let
        
            \begin{subequations}
                \begin{equation}
                    V^n = D^{k0}
                \end{equation}
                
                \begin{equation}
                    V^{n+c_i} = D^{ki}, \quad i=1,...,q
                \end{equation}
                
                \begin{equation}
                    V^{n+1} = D^{k(\gamma-1)}
                \end{equation}
            \end{subequations}
            
            where $D^{ki}$ are network prediction. General form of Runge-Kutta methods with q stages applied to Eq. (\ref{eq:system_eq}):
            
            \begin{subequations}
                \label{eq:rkm}
                \begin{equation}
                    V^{n+c_i} = V^{n} - \Delta t \sum_{j=1}^{q} a_{ij} F(V^{n+c_j}), \quad i=1,..,q
                \end{equation}
                
                \begin{equation}
                    V^{n+1} = V^{n} - \Delta t \sum_{j=1}^{q} b_{j} F(V^{n+c_j})
                \end{equation}
            \end{subequations}
                
            We use the Implicit Runge-Kutta methods with q stages and thus parameters $\{a_{ij}, b_j, c_j\}$ are chosen accordingly. Now, shifting second term on right hand side (RHS) in Eq. (\ref{eq:rkm}) to left hand side (LHS) and replacing exact operator $F$ with numerical gradient based operator $\hat{F}$
            \begin{subequations}
                \begin{equation}
                    \hat{W_i} = V^{n+c_i} + \Delta t \sum_{j=1}^{q} a_{ij} \hat{F}(V^{n+c_j}), \quad i=1,..,q
                \end{equation}
                
                \begin{equation}
                    \hat{W_{q+1}} = V^{n+1} + \Delta t \sum_{j=1}^{q} b_{j} \hat{F}(V^{n+c_j})
                \end{equation}
                
                \begin{equation}
                    P(\chi^k) = \sum_{i=1}^{q+1} \| vec( \hat{W_i} - V^n ) \|_2^2
                \end{equation}
            \end{subequations}
            where $\hat{W_i}$ are different estimates of $V^n$, $P(\chi^k)$ is the physics based loss function. For calculating loss the spatial gradients are approximated using Sobel filter 2D convolutions \cite{Sobel}. See $\S$ \ref{appA} for additional details. Note that physics-based loss function is only used during training of network.
            
        In next section, we present two examples to show model proposed is able to learn from sparse moving data labels. We illustrate the performance of the proposed approach with plots of prediction. We show that model is robust against noisy sensor measurements by showing error corresponding to various noise level. 
        Error used as a quantitative metric in plots in defined as
        
        \begin{equation}\label{eq:error}
            \epsilon^k = \frac{\left\|  D^{k \gamma^*}_{m} - J^{k*z+ \gamma^*}_{m}   \right\|_2}{\left\|   J^{k*z+ \gamma^*}_{m}  \right\|_2} \quad  k=0,...,\hat{N}
        \end{equation}
        
        where $\epsilon^k \in \mathbb R$ represents the error, $\hat{N} number of test samples $, $ J^{k*z+\gamma^*}_{m} \in \mathbb R^{\omega} $ are the true state and $ D^{k\gamma^*}_{m} \in \mathbb R^{\omega}$ are the predicted state using the proposed approach. $\left\| \cdot \right\|_2$ represents the L2 norm.

    \subsection{Experiment: 2D coupled Burgers’ equation}\label{sec:burgers}
        
        As the first example, we consider the the 2D coupled Burgers' system. It has the same convective and diffusion form as the in-compressible Navier-Stokes equations. It is an important model for understanding of various physical flows and problems, such as hydrodynamic turbulence, shock wave theory, wave processes in thermo-elastic medium, vorticity transport, dispersion in porous medium. The governing equations for Burgers' equation takes the following form:
        \begin{equation}
            \label{eq:b2}
            \bm u_t +\bm u \cdot \nabla \bm u - \nu \Delta \bm u = 0,
        \end{equation}
        
        with periodic boundary condition
        
        \begin{equation}\label{eq:bc_b2}
        \begin{split}
            \bm u \left(x=0, y, t \right) & =\bm u \left(x=L, y, t \right), \\
            \bm u \left(x, y=0, t \right) & =\bm u \left(x, y=L, t \right).
        \end{split}
        \end{equation}
        Eq. \eqref{eq:b2} can be written in expanded form as
        \begin{equation}
        \label{eq:b2m}
        \begin{aligned}
            \frac{\partial u}{\partial t} + u \frac{\partial u}{\partial x} + v \frac{\partial u}{\partial y} - \nu (\frac{\partial^2 u}{\partial x^2} + \frac{\partial^2 u}{\partial y^2}) = 0  \\
            \frac{\partial v}{\partial t} + u \frac{\partial v}{\partial x} + v \frac{\partial v}{\partial y} - \nu (\frac{\partial^2 v}{\partial x^2} + \frac{\partial^2 v}{\partial y^2}) = 0,
        \end{aligned}
        \end{equation}
        
        where $\nu$ is viscosity, $u$ and $v$ are the $x$ and $y$ components of velocity. We consider $\{x,y\} \in [0,1]$. The initial condition is defined using truncated Fourier series with random coefficients:
        
        \begin{equation}\label{eq:ic2}
           \bm  u(x, y, t=0) = \frac{2 \bm w(x, y)}{\max_{\{x, y\}} |\bm w(x, y)|} + \bm c,
        \end{equation}
        where
        \begin{equation}
           \bm w(x, y) = \sum_{i=-L}^{L}\sum_{j=-L}^{L} \bm a_{ij} \sin(2 \pi (ix+jy)) + \bm b_{ij} \cos(2 \pi(ix+jy)),
        \end{equation}
        
        where $\bm a_{ij}, \bm b_{ij} \sim \bm N(0, \mathbf I_2)$, $L=4$ and $\bm c \sim \bm {\mathcal U}(-1, 1) \in \mathbb R^2$.

        \subsubsection{Data-set and Model Parameters}
        
\begin{figure}[!htb]
                \centering
                \includegraphics[width=1\linewidth]{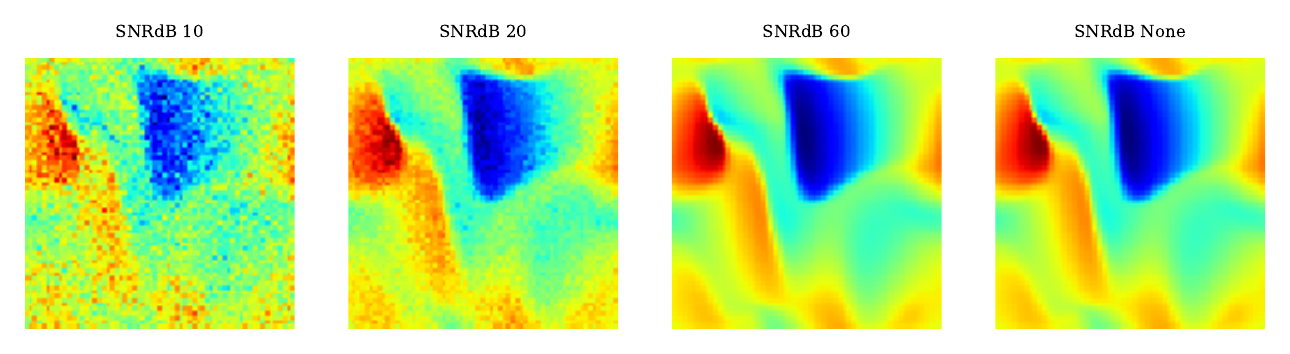}
                \caption{Plots of 2D Burgers velocity for visualizing different noise levels. SNRdB represents signal to noise ratio in decibels.}
                \label{fig:Burgers2DNoise}
            \end{figure}

            We use FeNICS \cite{LoggMardalEtAl2012} computing platform to solve the partial differential equations (\ref{eq:b2m}) for generating data-set. We discretize the spatial domain with $64 \times 64$ grid and use a time-step of $0.005$. Parameters related to data-set considered are displayed in Table \ref{tab:BurgersData}. We use noisy measurements to test the approach. Noise level is measured in signal to noise ratio (SNRdB) in decibels(dB) and is represented by $SNR_{dB}$. Signal after adding noise $r$ is formulated as:
        
            \begin{subequations}
                \begin{equation}
                    r = s + (\frac{P}{2 * 10^{\frac{SNR_{dB}}{10}}})^{0.5} * \mathcal N
                \end{equation}
                \begin{equation}
                    P = \frac{\sum s^2}{n}
                \end{equation}
            \end{subequations}
            where, $s$ is noise-free signal, $\mathcal N$ is random variable with standard normal distribution. Plots depicting different noise levels used for testing is shown in Fig. \ref{fig:Burgers2DNoise}.
            
            \begin{table}[!htb]
                \centering
                \caption{Data-set parameters for 2D Burgers problem.}
                \label{tab:BurgersData}
                
                \begin{tabular}{|p{1cm}|p{1cm}|p{1cm}|p{1cm}|p{1cm}|p{2cm}|}
                    \hline
                    
                    $\gamma$& $M$   & $L$   & $z$   & $\omega$  & domain        \\ \hline
                    5       & 2     & 50    & 2     & 3969      & $63\times63$  \\ \hline
                \end{tabular}
            \end{table}
        
            \begin{table}[!htb]
                \centering
                \caption{Network architecture of proposed Ensers for 2D coupled Burgers’ problem.}
                \label{tab:ta2_eg1}
                
                \begin{tabular}{|p{2cm}|p{2cm}|p{2cm}|p{2cm}|}
                    \hline
                    
                    Layer   & Input & Output            & Activation    \\ \hline
                    FC      & 8     & 64                & Softplus      \\ \hline
                    FC      & 64    & 64                & Softplus      \\ \hline
                    FC      & 64    & $\omega*M*\gamma$ & linear        \\ \hline
                    
                \end{tabular}
            \end{table}
        
            \begin{table}[!htb]
                \centering
                \caption{Training hyper parameters of Ensers for 2D Burgers problem.}
                \label{tab:BurgerstrainHP}
                
                \begin{tabular}{|p{1cm}|p{1cm}|p{1cm}|p{1cm}|p{1cm}|p{1cm}|p{1cm}|p{1cm}|p{1cm}|p{1cm}|p{1cm}|p{1cm}|}
                    \hline
                    
                    $\eta_o$& $\eta_{i0}$   & $\hat{\eta_{i}}$  & $\beta$   & $I_o$ & $I_i$ & $\zeta_0$ & $\hat{\zeta}$ & $p$   & $h$   & $N$ & $\varsigma$ \\ \hline
                    0.0002  & 0.1         & 0.006               & 11        & 2001  & 4     & 0.005     & 0.0001        & 32    & 800   & 22  & 8           \\ \hline
                \end{tabular}
            \end{table}
            
            \begin{table}[!htb]
                \centering
                \caption{Testing hyper-parameters of Ensers for 2D Burgers problem. $\eta_i$: inner learning rate, $I_i$: inner iterations}
                \label{tab:BurgerstestHP}
                
                \begin{tabular}{|p{1cm}|p{1cm}|p{1.5cm}|p{1cm}|}
                    \hline
                    
                    $\eta_i$  & $I_i$   & $p$       & $\hat{N}$ \\ \hline
                    5         & 100     & [4, 16]   & 12         \\ \hline
                \end{tabular}
            \end{table}
            
            FNN used in the model is a shallow network with three fully connected (FC) layers. We use Softplus \cite{glorot11a} activation function which has smooth first derivatives. Thus it helps avoid discontinuities because unrolling the inference procedure involves computing $\nabla_\theta \nabla_\xi (\mathfrak L)$. Network architecture is displayed in Table \ref{tab:ta2_eg1}. We use 800 data nodes for training from each system variable data which is $20\%$ of high-resolution data. Training hyper-parameters and training data are summarized in table \ref{tab:BurgerstrainHP}. For testing, we use sensor locations different from the ones used during training.
        
        \subsubsection{Results and Discussions}
        
            \begin{figure}[!htb]
                \centering
                \includegraphics[width=1\linewidth]{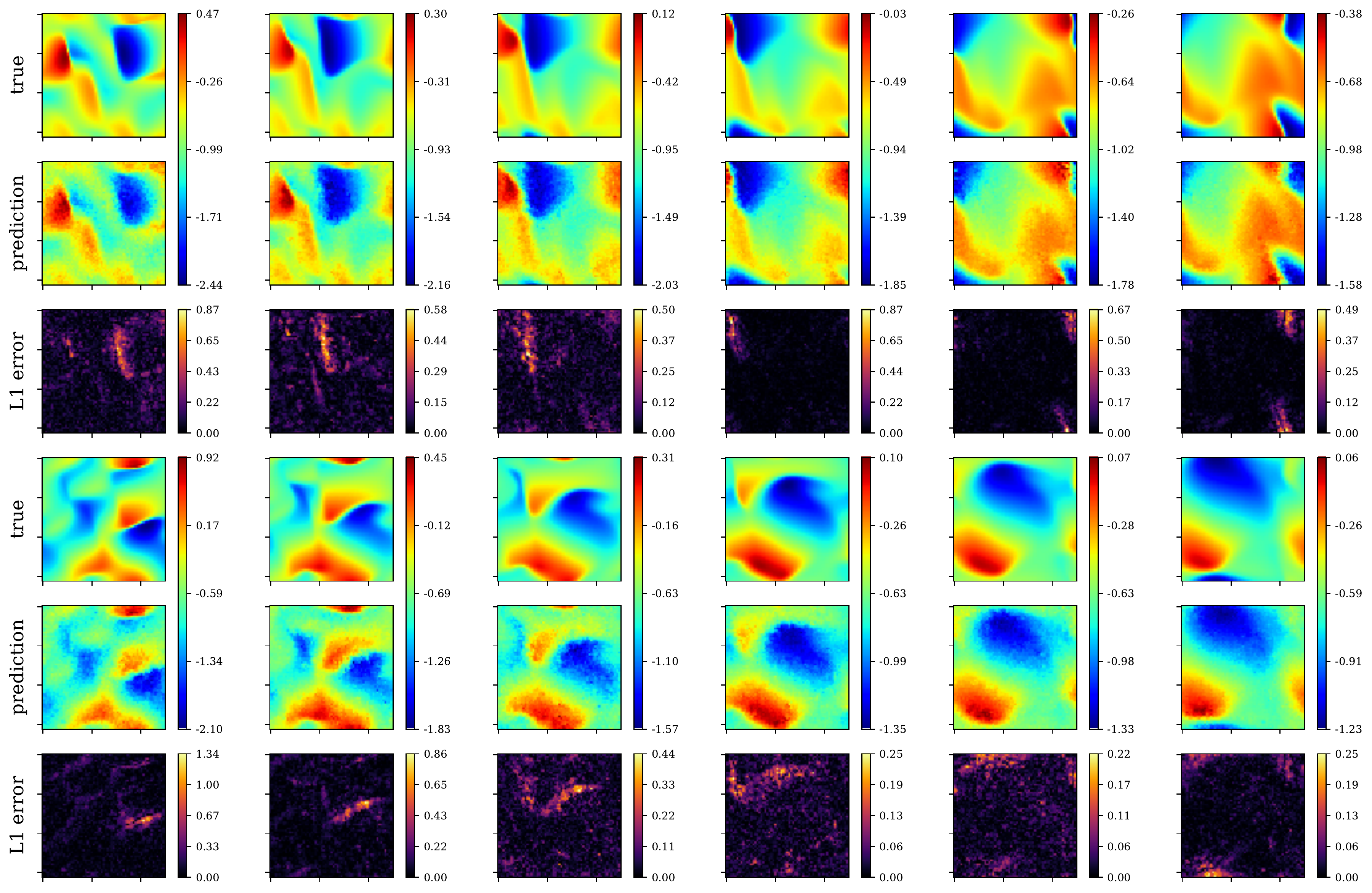}
                \caption{Ensers predictions of a 2D coupled Burgers’ test case. (Top to bottom) x-velocity FEM target solution, x-velocity Ensers prediction, x-velocity L1error, y-velocity FEM target solution, y-velocity Ensers prediction and y-velocity L1error.}
                \label{fig:fig1_eg1}
            \end{figure}
            
            \begin{figure}
                \centering
                \subfigure{ \includegraphics[width=0.45\textwidth]{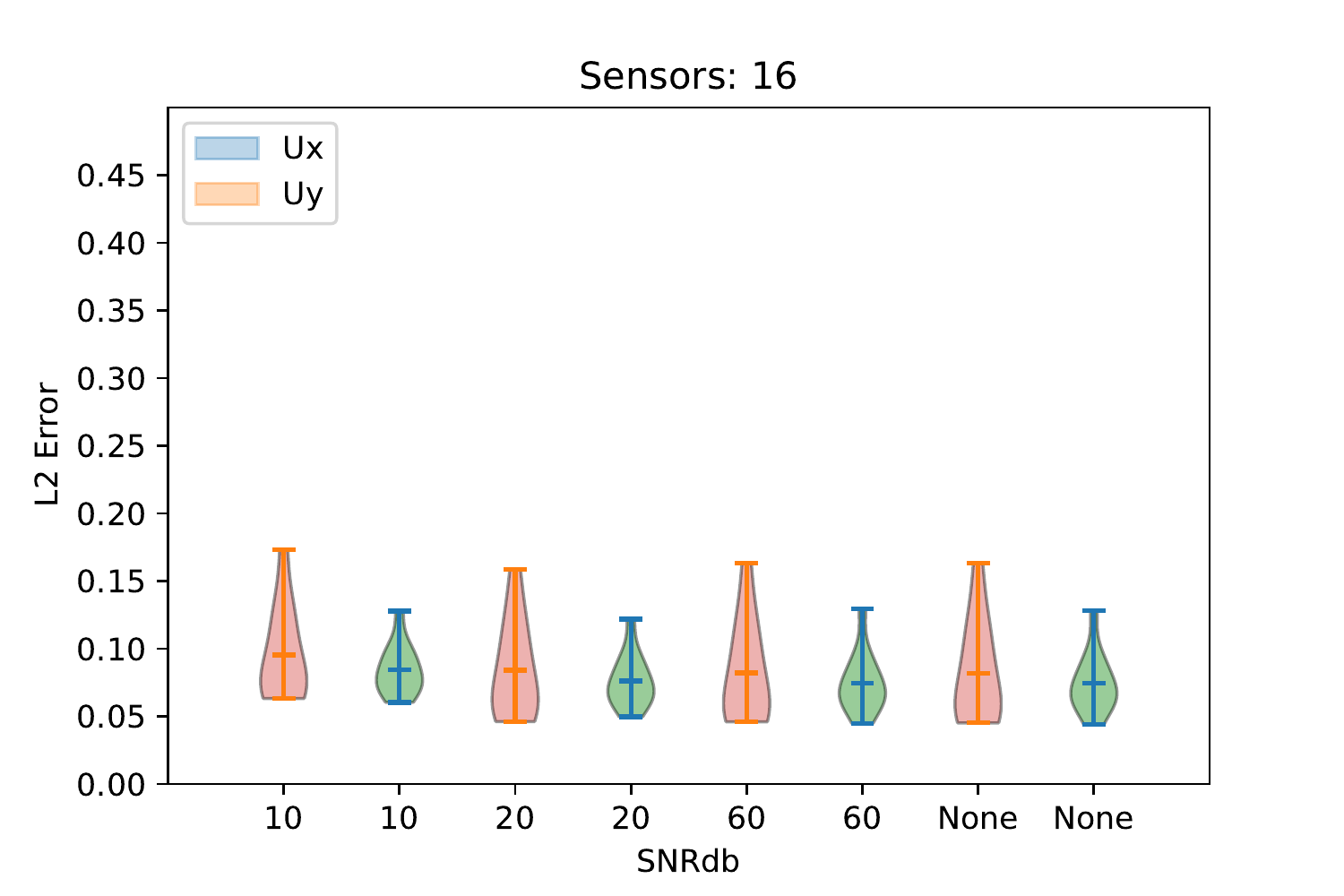}}
                \subfigure{ \includegraphics[width=0.45\textwidth]{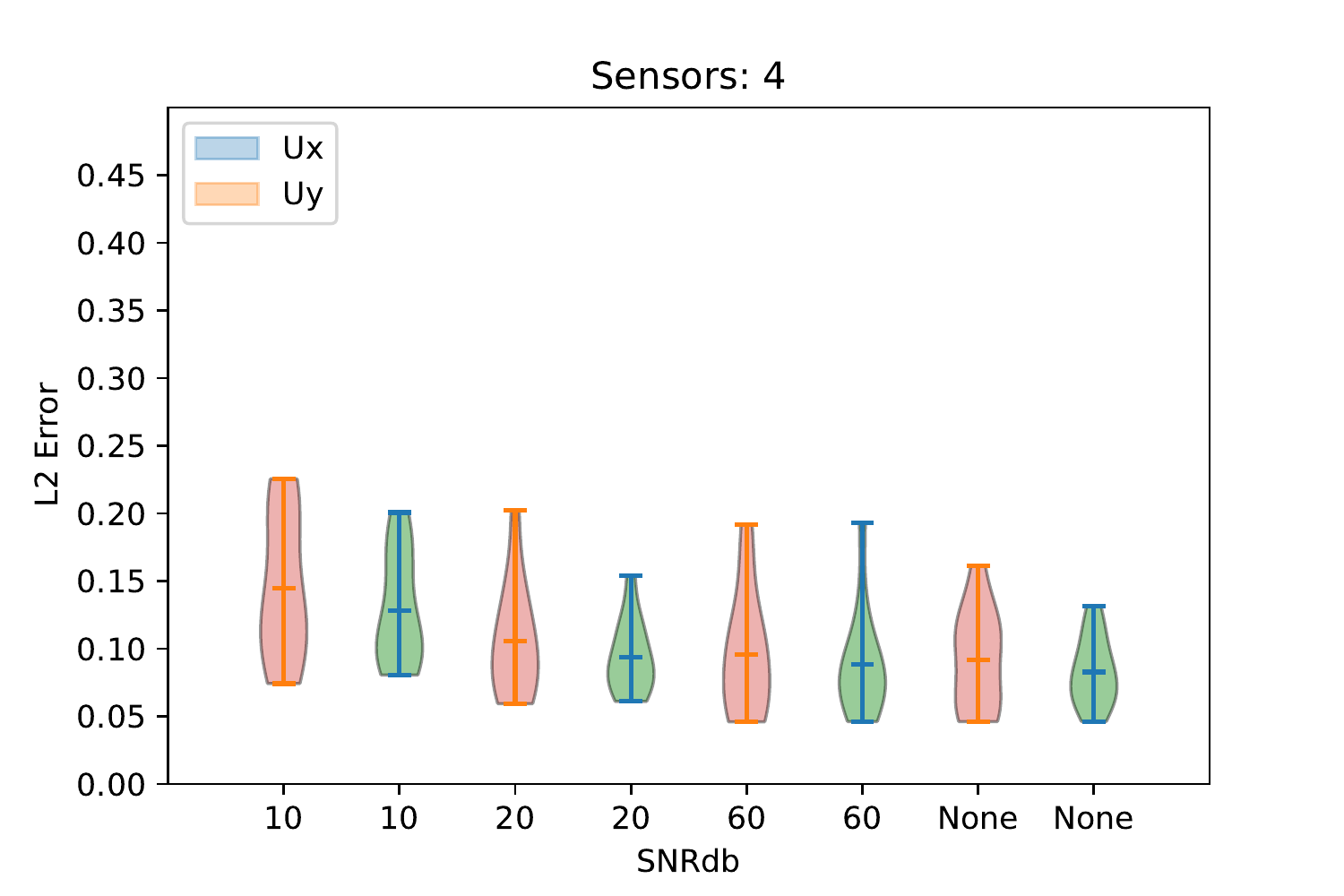}}
                
                \caption{Violin plot representing error $\epsilon$ distribution in prediction vector of x and y velocity for different noise levels in sensor measurements for 2D coupled Burgers’ test case.}
                \label{fig:Burgers2D_L2}
            \end{figure}
            
            Fig. \ref{fig:fig1_eg1} shows the prediction of Ensers for x and y velocity components at various times in simulation with 16 sensors. We see that it produces results that accurately captures the current state of the system at most points in the domain. If a network is trained without a physics-based loss function then it will produce garbage values at points where data is absent. Fig \ref{fig:Burgers2D_L2} shows violin plot of error vector $\epsilon$ defined in Eq. (\ref{eq:error}) between target and prediction with various noise level in sensor measurement for test case. In Eq. (\ref{eq:error}), $m$ for x velocity and y velocity are $0$ and $1$ respectively and $\gamma^*=2$. The plot represents the distribution of error $\epsilon$ in the prediction vector. Greater spread corresponding to a point on the y-axis corresponds to more values present in the error vector $\epsilon$ around that value of the point on the y-axis. We see the model is robust to noise as the mean error in the plot are close for cases of no noise($SNR_{dB}=None$) and high noise($SNR_{dB}=10$).

    \subsection{Experiment: Flow Past Cylinder}
    
        As the second example, we consider the flow past a cylinder problem. It is a well known canonical problem and is characterized by periodic laminar flow vortex shedding. System is governed by incompressible, laminar, Newtonian fluid equations:
        \begin{subequations}
            \label{eq:flow_eq}
            
            \begin{equation}
                \frac{\partial( u )}{\partial x} + \frac{\partial( v )}{\partial y} = 0
            \end{equation}
            
            \begin{equation}
                \frac{\partial( u )}{\partial t} + u \frac{\partial( u )}{\partial x} + v \frac{\partial( u )}{\partial y} = -\frac{1}{\rho} \frac{\partial( p )}{\partial x} + \nu \left( \frac{\partial^2( u )}{\partial x^2} + \frac{\partial^2( u )}{\partial y^2} \right)
            \end{equation}
            
            \begin{equation}
                \frac{\partial( v )}{\partial t} + u \frac{\partial( v )}{\partial x} + v \frac{\partial( v )}{\partial y} = -\frac{1}{\rho} \frac{\partial( p )}{\partial y} + \nu \left( \frac{\partial^2( v )}{\partial x^2} + \frac{\partial^2( v )}{\partial y^2} \right)
            \end{equation}
        \end{subequations}
        
        \subsubsection{Data-set and Model Parameters}
\begin{figure}
                \centering
                \subfigure[]{
                \includegraphics[width=0.5\textwidth]{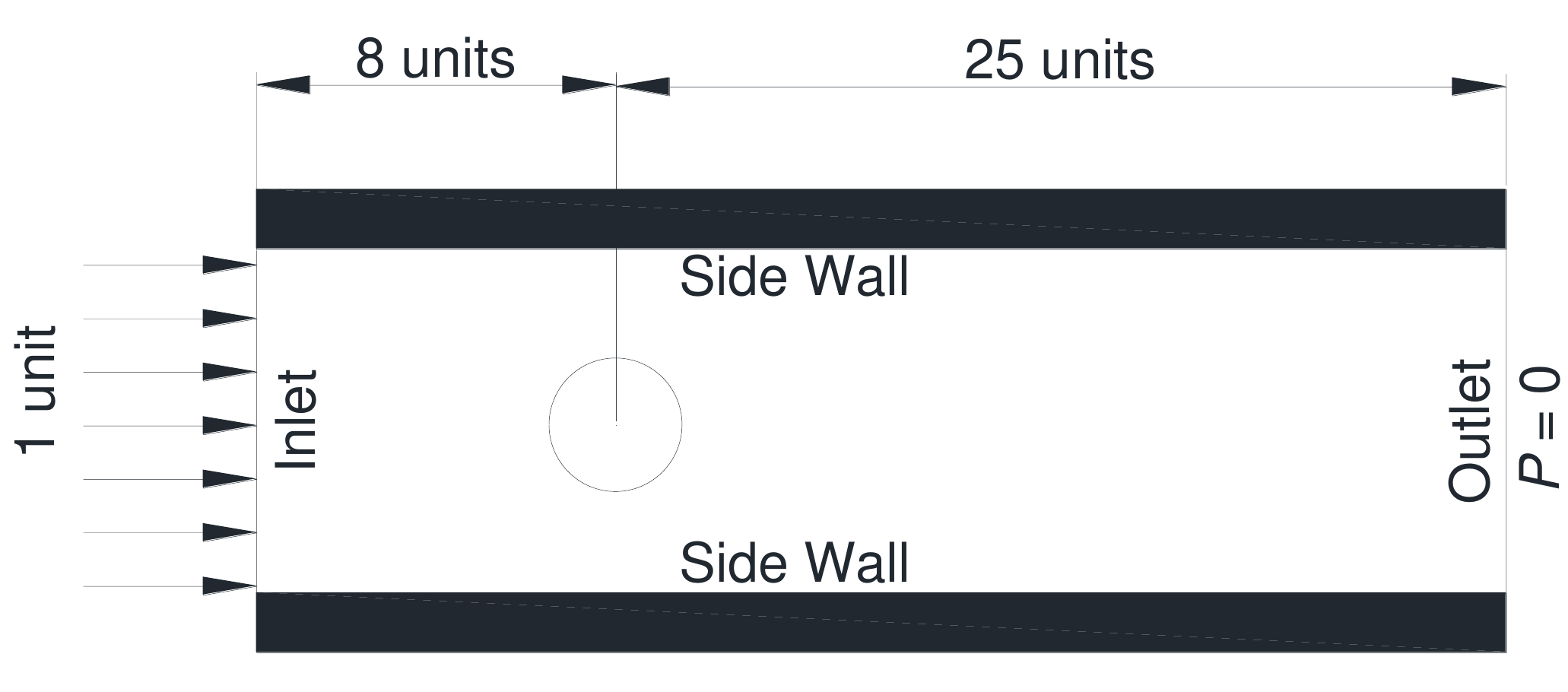}}
                \subfigure[]{
                \includegraphics[width=0.45\textwidth]{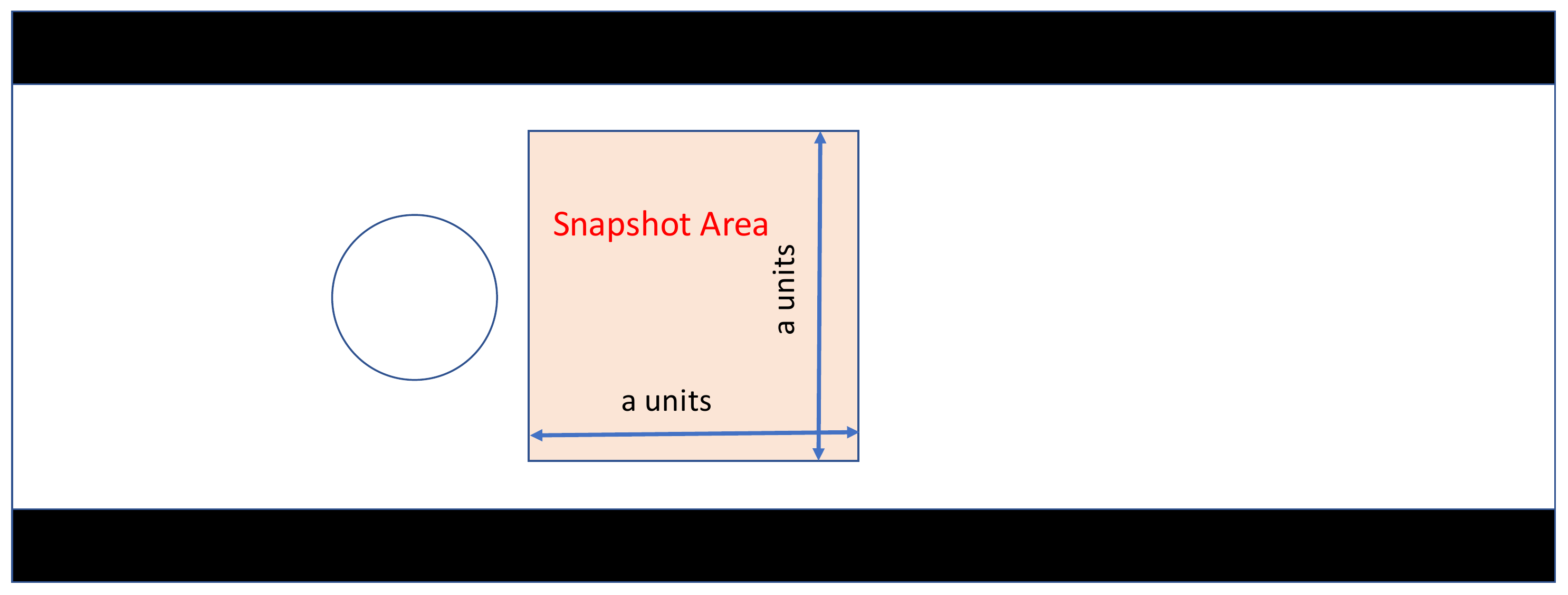}}
                \caption{(a) Schematic representation of the computational domain with boundary conditions at the inlet and the outlet. The cylinder has a diameter of 1 unit. A no-slip boundary is considered at the cylinder wall. Zero pressure gradient at the inlet and zero velocity gradient at the outlet are considered. (b) Schematic of the problem domain with snapshot cutout of $a \times a$. For flow past cylinder problems, $a = 4$ units. \it{The schematics are not to scale}.}
                \label{fig:domain}
            \end{figure}

            A schematic representation of the computational domain is shown in Fig. \ref{fig:domain}(a). The circular cylinder is considered to have a diameter of $1$ unit. The center of the cylinder is located at a distance of $8$ units from the inlet. The outlet is located at a distance of $25$ units from the center of the cylinder. The sidewalls are at $4$ units distance from the center of the cylinder. At the inlet boundary, a uniform velocity of $1$ unit along the $X$-direction is applied. Pressure boundary condition with $P=0$ is considered at the outlet. A no-slip boundary at the cylinder surface is considered. Coordinate of the snapshot cutout stretches from $[1.5,-2]\times [5.5, 2]$ which is discretized into $64 \times 64$ points in $x$ and $y$ directions (see Fig. \ref{fig:domain}(b)).
            
            \begin{figure}[!htb]
                \centering
                \includegraphics[width=1\linewidth]{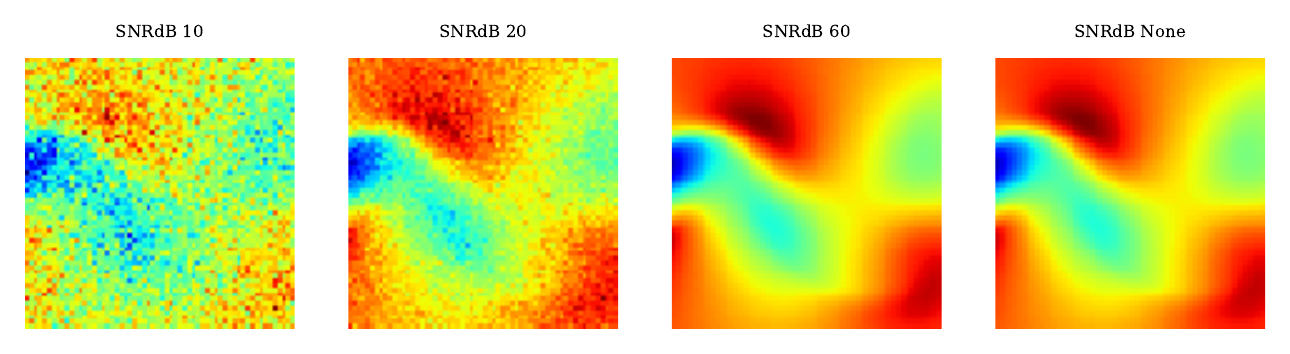}
                \caption{Plots of velocity of Flow Past Cylinder for visualizing different noise levels. SNRdB represents signal to noise ratio in decibels.}
                \label{fig:CylinderNoise}
            \end{figure}
            
            \begin{table}[!htb]
                \centering
                \caption{Data-set parameters for Flow Past Cylinder problem.}
                \label{tab:ta1_eg2}
                
                \begin{tabular}{|p{1cm}|p{1cm}|p{1cm}|p{1cm}|p{1cm}|p{1cm}|p{1cm}|p{2cm}|}
                    \hline
                    
                    $\gamma$& $M$   & $L$   & $z$   & $\omega$  & $\nu$ & Re    & domain        \\ \hline
                    5       & 3     & 25    & 1     & 4096      & 0.005 & 200   & $64\times64$  \\ \hline
                \end{tabular}
            \end{table}
            
            The data-set is generated by using Unsteady Reynolds-averaged Navier Stokes (URANS) simulation in OpenFoam \citep{jasak2009openfoam}. The overall problem domain is discretized into 63420 elements with finer mesh near the cylinder. Time step $\delta t = 0.02$ units is considered. Code for OpenFOAM simulation can be found at \cite{flowOpenFoam}. Parameters related to the data-set considered are displayed in Table \ref{tab:ta1_eg2}.
        
            We use a shallow network in the model with two fully connected(FC) layers. Network architecture is displayed in Table \ref{tab:ta2_eg2}. We use 500 data nodes for training from each system variable data which is $12.2\%$ of high-resolution data. Training and testing hyper-parameters are shown in table \ref{tab:FPCTrainHP} and \ref{tab:FPCTestHP} respectively. Sensor locations for testing are different from the ones used during training. For this problem physics-based loss function is extended to include the continuity equation:
            
            \begin{equation}
                    P(\chi^k) = \sum_{i=1}^{q+1} \| vec( W_i - V^n ) \|_2^2 + \sum_{i=0}^{\gamma-1} \| vec(\frac{\partial( V^n_0 )}{\partial x} + \frac{\partial( V^n_1 )}{\partial y}) \|_2^2
            \end{equation}

            \begin{table}[!htb]
                \centering
                \caption{Network architecture of proposed Ensers for Flow Past Cylinder problem.}
                \label{tab:ta2_eg2}
                
                \begin{tabular}{|p{2cm}|p{2cm}|p{2cm}|p{2cm}|}
                    \hline
                    
                    Layer   & Input & Output            & Activation    \\ \hline
                    FC      & 8     & 64                & Softplus      \\ \hline
                    FC      & 64    & $\omega*M*\gamma$ & linear        \\ \hline
                    
                \end{tabular}
            \end{table}
        
            \begin{table}[!htb]
                \centering
                \caption{Training hyper parameters of Ensers for Flow Past Cylinder.}
                \label{tab:FPCTrainHP}
                
                \begin{tabular}{|p{1cm}|p{1cm}|p{1cm}|p{1cm}|p{1cm}|p{1cm}|p{1cm}|p{1cm}|p{1cm}|p{1cm}|p{1cm}|p{1cm}|}
                    \hline
                    
                    $\eta_o$& $\eta_{i0}$   & $\hat{\eta_{i}}$  & $\beta$   & $I_o$ & $I_i$ & $\zeta_0$ & $\hat{\zeta}$ & $p$   & $h$   & $N$ & $\varsigma$   \\ \hline
                    0.0003  & 0.1           & 0.002             & 9         & 3001  & 5     & 0.01      & 0.0006        & 16    & 500   & 18  & 8   \\ \hline
                \end{tabular}
            \end{table}
            
            \begin{table}[!htb]
                \centering
                \caption{Testing hyper parameters of Ensers for Flow Past Cylinder problem. $\eta_i$: inner learning rate, $I_i$: inner iterations}
                \label{tab:FPCTestHP}
                
                \begin{tabular}{|p{1cm}|p{1cm}|p{1.5cm}|p{1cm}|}
                    \hline
                    
                    $\eta_i$  & $I_i$   & $p$       & $\hat{N}$ \\ \hline
                    5         & 100     & [4, 16]   & 12         \\ \hline
                \end{tabular}
            \end{table}
        
        \subsubsection{Results and Discussions}
            \begin{figure}[!htb]
                \centering
                \includegraphics[width=1\linewidth]{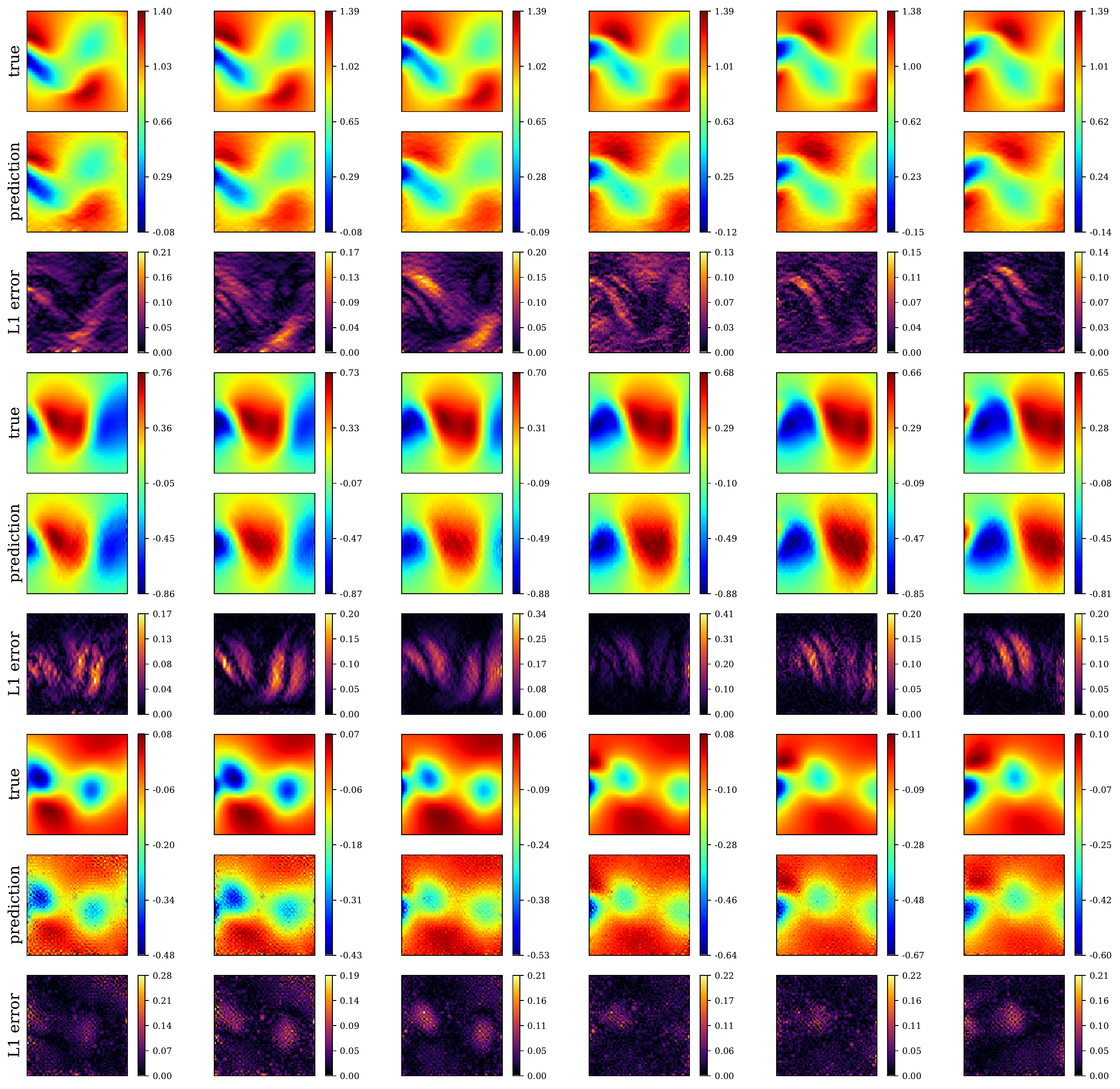}
                \caption{Ensers predictions of a Flow Past Cylinder test case. (Top to bottom) x-velocity target solution, x-velocity Ensers prediction, x-velocity L1error, y-velocity target solution, y-velocity Ensers prediction, y-velocity L1error, pressure target solution, pressure Ensers prediction and pressure L1error.}
                \label{fig:fig1_eg2}
            \end{figure}
            
            \begin{figure}
                \centering
                \subfigure[]{ \includegraphics[width=0.45\textwidth]{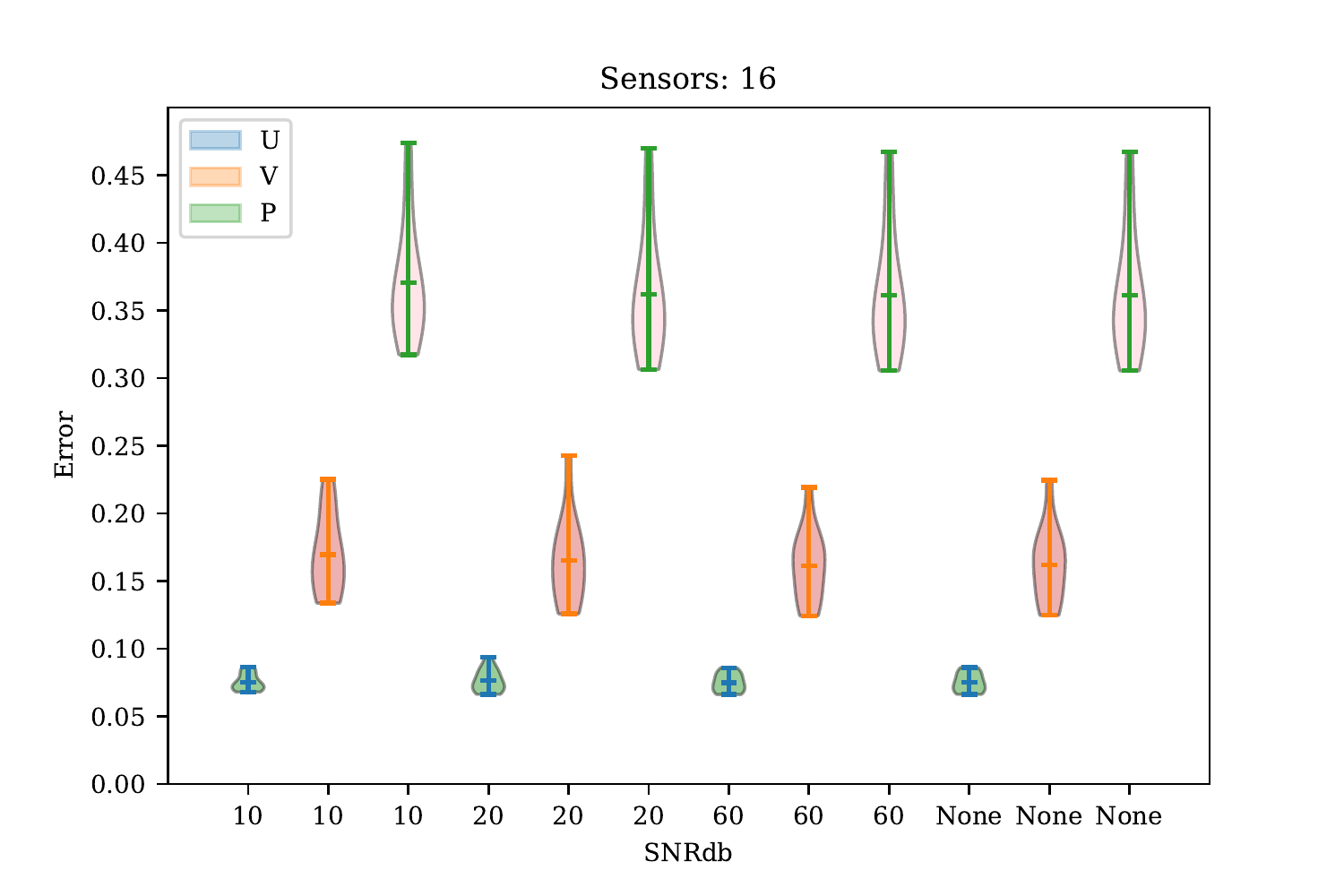}}
                \subfigure[]{ \includegraphics[width=0.45\textwidth]{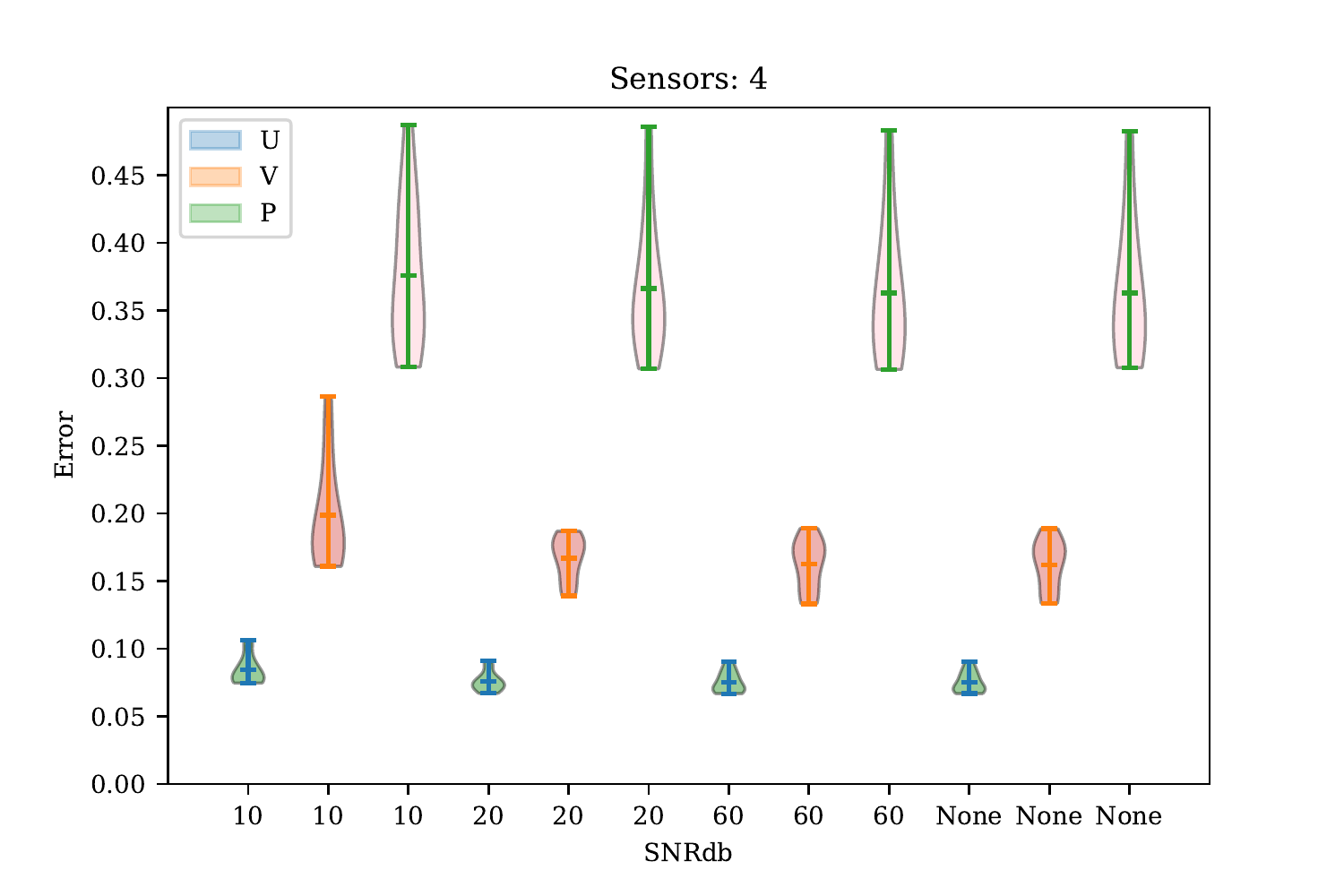}}
                
                \caption{Violin plot representing error $\epsilon$ distribution in prediction vector of x,y velocity and pressure for different noise levels in sensor measurements for Flow Past Cylinder test case.}
                \label{fig:Cylinder_L2}
            \end{figure}
            
            Fig. \ref{fig:fig1_eg2} shows prediction of Ensers for pressure, x and y velocity components with $16$ sensors. We see that it produces results that accurately capture the current state of the system at most points in the domain. Fig. \ref{fig:Cylinder_L2} shows violin plot of error defined in Eq. (\ref{eq:error}) between target and prediction with various noise levels in sensor measurement for the test case. In Eq. (\ref{eq:error}), $m$ for x velocity, y velocity and pressure are $0$, $1$ and $2$ respectively and $\gamma^*=2$. The plot represents the distribution of error in prediction vectors. We see model is robust to noise as mean error in Fig \ref{fig:Cylinder_L2} are close for cases of no noise($SNR_{dB}=None$) and high noise($SNR_{dB}=10$).

\section{Continuous space models}\label{sec:csm}

    \subsection{Proposed approach}\label{sec:paC}
         In this section, we propose a novel deep learning-based continuous framework for state estimation. Prediction of $\gamma$ high dimensional states is done via multiple passes from a feed-forward neural network (FNN) $\Gamma$ for every collocation point with coordinate vector $X_r$ using optimized reduced state vectors $\xi^k \in \mathbb R^{\varsigma}$:
    
        \begin{equation}
            \label{eq:decoderC}
            \begin{split}
            \Gamma(\Xi^k_r) = \{ D^{ki}_{mr} \in \mathbb R \ | \ i=0,...,\gamma-1, \ m=0,...,M-1 \}, \quad \ r=0,...,\omega
            \end{split}
        \end{equation}
        
        \begin{equation}
            \label{eq:eX}
            \Xi^k_r = \{ \xi^k, X_r \}, \quad \ r=0,...,\omega
        \end{equation}
        where output vector of FNN $\Gamma$ has dimension $\gamma*M$, $D^{ki}_{mr}$ is predicted states value at $r^{th}$ collocation point, $X_r$ are coordinates of $r^{th}$ collocation point. Output of $\omega$ FNNs are combined to form third-order tensor $D^k \in \mathbb R^{ \gamma \times M \times \omega} $. Reduced state is obtained by solving the following minimization problem using sensor data $\chi^k$ and predicted states by the neural network.
        
        \begin{subequations}
            \label{eq:optimizeC}
            
            \begin{equation}
                \xi^k = \arg \min_{ \tilde{ \xi^k } } \| vec(\chi^k - \rho( \tilde{ \Xi^k } )) \|_2^2
            \end{equation}
            
            \begin{equation}
                \rho(\tilde{ \Xi^k }) = \{ \tilde{Q}^{ki}_m \in \mathbb R^{p} \ | \ i=0,...,\gamma-1, \ m=0,...,M-1 \}
            \end{equation}
            
            \begin{equation}
                \label{eq:ValSenser_C}
                \tilde{Q}^{ki}_m = \Lambda^{kz+i}_m \tilde{D}^{ki}_m
            \end{equation}
            
            \begin{equation}
                \label{eq:NN_Xi_tilde_C}
            \begin{split}
                \Gamma(\tilde{ \Xi^k_r }) = \{ \tilde{D}^{ki}_{mr} \in \mathbb R \ | \ i=0,...,\gamma-1, \ m=0,...,M-1 \}, \\ \quad r=0,...,\omega
            \end{split}
            \end{equation}
            
            \begin{equation}
                \tilde{ \Xi^k_r } = \{ \tilde{ \xi^k }, X_r \}, \quad r=0,...,\omega
            \end{equation}
        \end{subequations}
        where $\tilde{Q}^{ki}_m \in \mathbb R^{p}$ are values of predicted states at sensor locations $\lambda^{kz+i}_m$. Note that in practice $\xi^k$ is obtained by a few steps of gradient descent instead of global minimization. The network is trained by minimizing data loss and physics-based loss $P(\chi^k)$ across training samples $N$.
        
        \begin{subequations}
            \begin{equation}
                \theta^* = \arg \min_{ \theta } \sum_{k=0}^{N-1} \| vec(\phi^k - \Upsilon( \chi^k )) \|_2^2 + P(\chi^k)
            \end{equation}
            
            \begin{equation}
                \label{eq:train_bC}
                \Upsilon( \chi^k ) = \{ Y^{ki}_m \in \mathbb R^{h} \ | \ i=0,...,\gamma-1, \ m=0,...,M-1 \}
            \end{equation}
            
            \begin{equation}
                \label{eq:ValC}
                Y^{ki}_m = \Pi^{kz+i}_m D^{ki}_m
            \end{equation}
        \end{subequations}
        where $D^{ki}_m \in \mathbb R^{\omega}$ are predicted states from Eq. (\ref{eq:decoderC}), $Y^{ki}_m \in \mathbb R^{h}$ are values of predicted states at data locations $\pi^{kz+i}_m$, $\Upsilon( \chi^k ) \in \mathbb R^{\gamma \times M \times p}$ are predicted states composed of $\gamma$ time steps, $\theta$ are network parameters. Fig. \ref{fig:architecture} shows the network architecture during training. Equations (\ref{eq:decoderC}) and (\ref{eq:optimizeC}) together forms the implicit optimization layer shown in the Fig. \ref{fig:architecture} for the continuous model. Training and testing procedure is shown in Algorithm \ref{alg:trainC} and \ref{alg:testC} respectively.

        \begin{algorithm}[!htb]
        
            \caption{Training ENSERS}
            \label{alg:trainC}
            \begin{algorithmic}[1]
        
            \STATE{\textbf{Inputs:} $Z, S, T$.     \COMMENT{Eq. \eqref{eq:simData}, Eq. \eqref{eq:senLocEachT}, Eq. \eqref{eq:dataLocEachT}} }
            
            \STATE{\textbf{Set Hyper-parameters:} $\eta_o$: outer learning rate, $\eta_{i0}$: inner learning rate at epoch=0, $\hat{\eta_{i}}$: inner learning rate rate, $\beta$: batch size, $I_o$: outer iterations, $I_i$: inner iterations, $\hat{\zeta}$: physics penalty rate, $\zeta_0$: physics penalty at epoch=0, $N$: number of train samples, $z$, $\gamma$.}
            
            \STATE{\textbf{Calculate data-set:} $\chi$, $\phi$, $\Lambda$, $\Pi$}
            
            \STATE{\textbf{Initialize:} Neural network model: $ \Gamma(\cdot; \theta)$}
        
            \FOR{ $\iota_o = 0$ \textbf{to} $I_o-1$}        \COMMENT{Outer optimization Loop}
            
                \STATE{$\eta_{i} = \eta_{i0} + \iota_o \hat{\eta_{i}}$} \COMMENT{Schedule inner learning rate}
                \STATE{$\zeta = \zeta_0 + \iota_o \hat{\zeta}$}         \COMMENT{Schedule physics penalty}
                \FOR{$k = 0$ \textbf{to} $N-1$}
                
                    \STATE{\textbf{Initialize:} $\tilde{ \xi^k }$}
                    \FOR{$\iota_i = 0$ \textbf{to} $I_i-1$}        \COMMENT{Inner optimization Loop}
                    
                        \STATE{$\tilde{ \Xi^k_r } = \{ \tilde{ \xi^k }, X_r \}, \quad r=0,...,\omega$ }
                        
                        \STATE{$ \tilde{D^k_r} = \Gamma( \tilde{ \Xi^k_r } ), \quad r=0,...,\omega $}   \COMMENT{Eq. \eqref{eq:NN_Xi_tilde_C}}
                        \STATE{$ \tilde{Q^k} \leftarrow$ DrawValuesAtSensorLocations($\tilde{D^k}$) }   \COMMENT{Eq. \eqref{eq:ValSenser_C}}
                        \STATE{$\mathfrak L = MSE(  \chi^k, \tilde{Q^k} )$}
                        \STATE{$\frac{\partial \mathfrak L}{\partial \tilde{ \xi^k }} \leftarrow$ Backprop($\mathfrak L$)}
                        \STATE{$ \tilde{ \xi^k } = \tilde{ \xi^k } -  \eta_i \frac{\partial \mathfrak L}{\partial \tilde{ \xi^k }}$ }
                    \ENDFOR
                    
                    \STATE{$\hat{X_r} = X_r$   }
                    \STATE{$\Xi^k_r = \{ \tilde{ \xi^k }, \hat{X_r} \}, \quad r=0,...,\omega$ }
                    \STATE{$D^k_r = \Gamma(\Xi^k_r), \quad r=0,...,\omega$   \COMMENT{Eq. \eqref{eq:decoderC}}}
                    \STATE{$Y^k \leftarrow$ DrawValuesAtDataLocations($D^k$)    \COMMENT{Eq. \eqref{eq:ValC}}    }
                    \STATE{$\mathcal L = MSE(\phi^k, Y^k) + \zeta P(D^k)     $\COMMENT{Calculate loss}}
                    \STATE{$\frac{\partial \mathcal L}{\partial \theta} \leftarrow$ Backprop($\mathcal L$) }
                    \STATE{$ \theta \leftarrow \theta - \eta_o \frac{\partial \mathcal L}{\partial \theta}$        \COMMENT{Update weights} }
                \ENDFOR
            \ENDFOR
        
            \STATE{\textbf{Output:} Trained network $ \Gamma(\cdot; \theta^*)$.}
        
        \end{algorithmic}
        \end{algorithm}
        
        \begin{algorithm}[ht]
            \caption{Testing ENSERS}
            \label{alg:testC}
            \begin{algorithmic}[1]
        
            \STATE{\textbf{Inputs:} Trained network $ \Gamma(\cdot; \theta^*)$, $S$.     \COMMENT{Eq. \eqref{eq:senLocEachT}} }
            
            \STATE{\textbf{Set Hyper-parameters:} $\eta_i$: inner learning rate, $I_i$: inner iterations, $\hat{N}$: number of test samples, $z$, $\gamma$.}
            
            \STATE{\textbf{Calculate data-set:} $\chi$, $\Lambda$}
            
            \FOR{$k = 0$ \textbf{to} $\hat{N}-1$}
            
                \STATE{\textbf{Initialize:} $\tilde{ \xi^k }$  }
                \FOR{$\iota_i = 0$ \textbf{to} $I_i-1$}        \COMMENT{Inner optimization Loop}
                    
                    \STATE{$\tilde{ \Xi^k } = \{ \tilde{ \xi^k }, X_r \}, \quad r=0,...,\omega$ }
                    
                    \STATE{$ \tilde{D^k_r} = \Gamma( \tilde{ \Xi^k_r } ), \quad r=0,...,\omega $}   \COMMENT{Eq. \eqref{eq:NN_Xi_tilde_C}}
                    \STATE{$ \tilde{Q^k} \leftarrow$ DrawValuesAtSensorLocations($\tilde{D^k}$) }   \COMMENT{Eq. \eqref{eq:ValSenser_C}}
                    \STATE{$\mathfrak L = MSE(  \chi^k, \tilde{Q^k} )$}
                    \STATE{$\frac{\partial \mathfrak L}{\partial \tilde{ \xi^k }} \leftarrow$ Backprop($\mathfrak L$) }
                    \STATE{$\tilde{ \xi^k } = \tilde{ \xi^k } -  \eta_i \frac{\partial \mathfrak L}{\partial \tilde{ \xi^k }} $}
                \ENDFOR
                
                \STATE{$\Xi^k_r = \{ \tilde{ \xi^k }, X_r \}, \quad r=0,...,\omega$ }
                \STATE{$D^k_r = \Gamma(\Xi^k_r), \quad r=0,...,\omega$   \COMMENT{Eq. \eqref{eq:decoderC}} }
            \ENDFOR
             
            \STATE{\textbf{Output:} Predicted states $\{ D^{ki}_m \ | \ k=0,...,N-1, \ i=0,...,\gamma-1, \ m=0,...,M-1 \}$.}
        
        \end{algorithmic}
        \end{algorithm}
        
        \subsubsection{Physics-based loss function}
            
            The physics-based loss function for continuous space models differs from discrete formulation due to how RHS of Eq. (\ref{eq:system_eq}) i.e. $F$ is evaluated. In this case, we use automatic differentiation for calculating gradients w.r.t. coordinates used in $F$. For example in first case of  Allen–Cahn equation $\hat{F}$ is evaluated as:
            
            \begin{equation}
                \label{eq:ACeqAutoDiff}
                \hat{F_r} = 0.0001 \frac{\partial^2 u}{\partial \hat{X_r}^2} + 5u^3 - 5u, \quad r=0,...,\omega
            \end{equation}
            
            where $\hat{X_r}$ is coordinate vector concatenated with optimized reduced state vector at end of inner optimization loop, see line $18$ in algorithm (\ref{alg:trainC}). In second case of Convection-diffusion equation $\hat{F}$ is evaluated as:
            
            \begin{equation}
                \label{eq:CDeqAutoDiff}
                \hat{F_r} = a(x, y) \frac{\partial u}{\partial \hat{X_r^0}} + b(x, y) \frac{\partial u}{\partial \hat{X_r^1}} + c \frac{\partial^2 u}{\partial \hat{X_r^0}^2} + d \frac{\partial^2 u}{\partial \hat{X_r^1}^2}, \quad r=0,...,\omega.
            \end{equation}            

            where $\hat{X_r^0}, \hat{X_r^1}$ are x and y coordinate respectively, $a, b, c, d$ are defined in Eq. \ref{eq:CDcoeff}.

    \subsection{Experiment: Allen–Cahn equation}\label{sec:AllenCahn}
    
        We consider the Allen–Cahn equation along with periodic boundary conditions. The Allen–Cahn equation is a well-known equation from the area of reaction-diffusion systems. It describes the process of phase separation in multicomponent alloy systems, including order-disorder transitions. 
        
        \begin{equation}
            \label{eq:ACeq}
            \begin{aligned}
            u_t - 0.0001u_{xx} + 5u^3 - 5u = 0, \quad x \in [-1, 1], t \in [0, 1], \\
            u(0, x) = x^2 cos(\pi x), \\
            u(t, -1) = u(t, 1), \\
            u_x(t, -1) = u_x(t, 1).
            \end{aligned}
        \end{equation}
        
        \subsubsection{Data-set and Model Parameters}
        
\begin{figure}[!htb]
                \centering
                \includegraphics[width=1\linewidth]{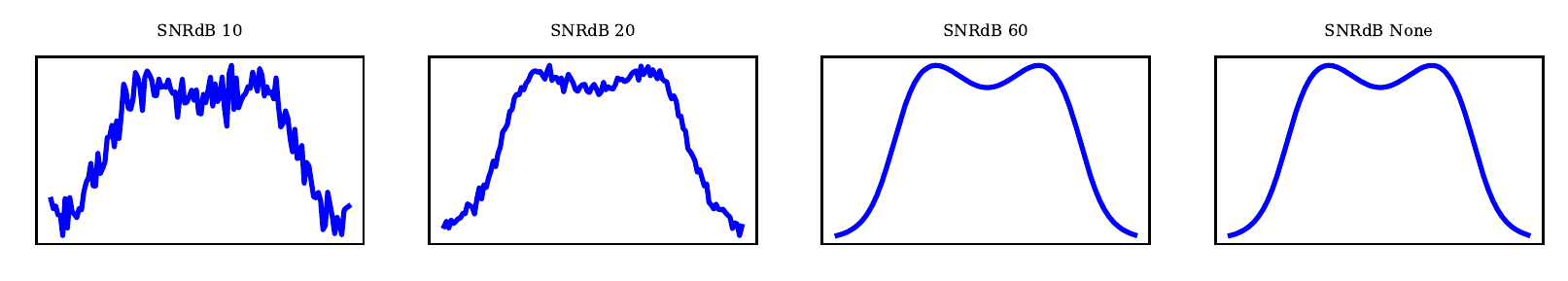}
                \caption{Data plots of Allen–Cahn equation for visualizing different noise levels. SNRdB represents signal to noise ratio in decibels.}
                \label{fig:ACNoise}
            \end{figure}

            Data-set is generated by simulating the Allen–Cahn equation (\ref{eq:ACeq}) using conventional spectral methods. Starting from an initial condition $u(0, x) =x^2 cos(\pi x)$ and assuming periodic boundary conditions $u(t, -1) = u(t, 1)$ and $u_x(t, -1) = u_x(t, 1)$, we integrated Eq. (\ref{eq:ACeq}) up to a final time $t=1.0$ using the Chebfun package \cite{driscoll2014chebfun} with a spectral Fourier discretization with 512 modes and a fourth-order explicit Runge–Kutta temporal integrator with time-step $\Delta t=10^{-5}$. For more details on the data-set see \cite{RAISSI_p}. Plots depicting different noise levels used for sensor measurement during testing are shown in Fig. \ref{fig:ACNoise}.
            
            \begin{table}[!htb]
                \centering
                \caption{Data-set parameters for Allen–Cahn equation.}
                \label{tab:ta1_eg1}
                
                \begin{tabular}{|p{1cm}|p{1cm}|p{1cm}|p{1cm}|p{1cm}|p{2cm}|}
                    \hline
                    
                    $\gamma$& $M$   & $L$   & $z$   & $\omega$  & domain        \\ \hline
                    5       & 1     & 50    & 1     & 128      & $128$  \\ \hline
                \end{tabular}
            \end{table}
            
            The network considered in the continuous formulation has significantly fewer weights compared to the discrete one. This is because the output of the network is predicted at a single point. Network architecture is shown in table \ref{tab:ACnet}. The input size of the network is equal to the sum of the size of the reduced state $\varsigma$ and the dimension of coordinates. For training we use 10 data nodes $p$ which is $7.81\%$ of $128$ total nodes $\omega$. Similar to previous cases inner loop learning rate $\eta_i$ and physics loss penalty $\zeta$ are increased linearly with each epoch. Other training and testing hyperparameters are shown in table \ref{tab:ACTrainHP} and \ref{tab:ACTestHP} respectively.
        
            \begin{table}[!htb]
                \centering
                \caption{Network architecture of proposed Ensers for Allen–Cahn equation.}
                \label{tab:ACnet}
                
                \begin{tabular}{|p{2cm}|p{2cm}|p{2cm}|p{2cm}|}
                    \hline
                    
                    Layer   & Input & Output            & Activation    \\ \hline
                    FC      & 7    & 128                & Tanh          \\ \hline
                    FC      & 128  & 128                & Tanh          \\ \hline
                    FC      & 128  & 128                & Tanh          \\ \hline
                    FC      & 128  & 128                & Tanh          \\ \hline
                    FC      & 128  & $M*\gamma$         & linear        \\ \hline
                    
                \end{tabular}
            \end{table}
        
            \begin{table}[!htb]
                \centering
                \caption{Training hyper parameters of Ensers for Allen–Cahn equation.}
                \label{tab:ACTrainHP}
                
                \begin{tabular}{|p{1cm}|p{1cm}|p{1cm}|p{1cm}|p{1cm}|p{1cm}|p{1cm}|p{1cm}|p{1cm}|p{1cm}|p{1cm}|p{1cm}|}
                    \hline
                    
                    $\eta_o$& $\eta_{i0}$   & $\hat{\eta_{i}}$  & $\beta$   & $I_o$ & $I_i$ & $\zeta_0$ & $\hat{\zeta}$     & $p$   & $h$   & $N$ & $\varsigma$   \\ \hline
                    0.001   & 0.005         & $6\mathrm{e}{-5}$ & 10        & 1201  & 8     & 0.01      & $2\mathrm{e}{-5}$ & 10    & 10   & 40  & 6             \\ \hline
                \end{tabular}
            \end{table}
            
            \begin{table}[!htb]
                \centering
                \caption{Testing hyper parameters of Ensers for Allen–Cahn equation. $\eta_i$: inner learning rate, $I_i$: inner iterations}
                \label{tab:ACTestHP}
                
                \begin{tabular}{|p{1cm}|p{1cm}|p{1.5cm}|p{1cm}|}
                    \hline
                    
                    $\eta_i$    & $I_i$ & $p$       & $\hat{N}$ \\ \hline
                    0.1         & 50    & [6, 16]   & 12         \\ \hline
                \end{tabular}
            \end{table}

        \subsubsection{Results and Discussions}
        
\begin{figure}[!htb]
                \centering
                \includegraphics[width=1\linewidth]{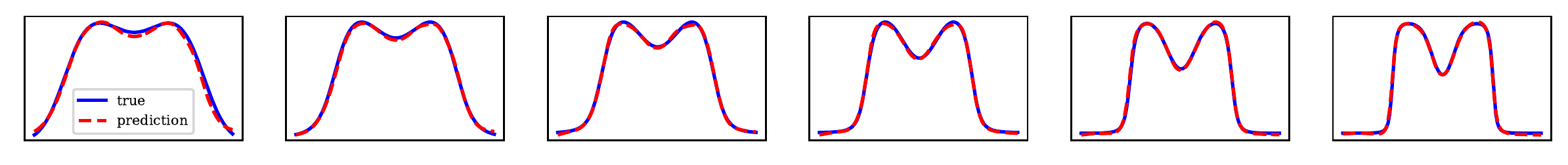}
                \caption{Ensers predictions of a Allen–Cahn equation test case with 16 sensors and no noise.}
                \label{fig:AC_plot}
            \end{figure}

\begin{figure}
                \centering
                \subfigure[]{ \includegraphics[width=0.45\textwidth]{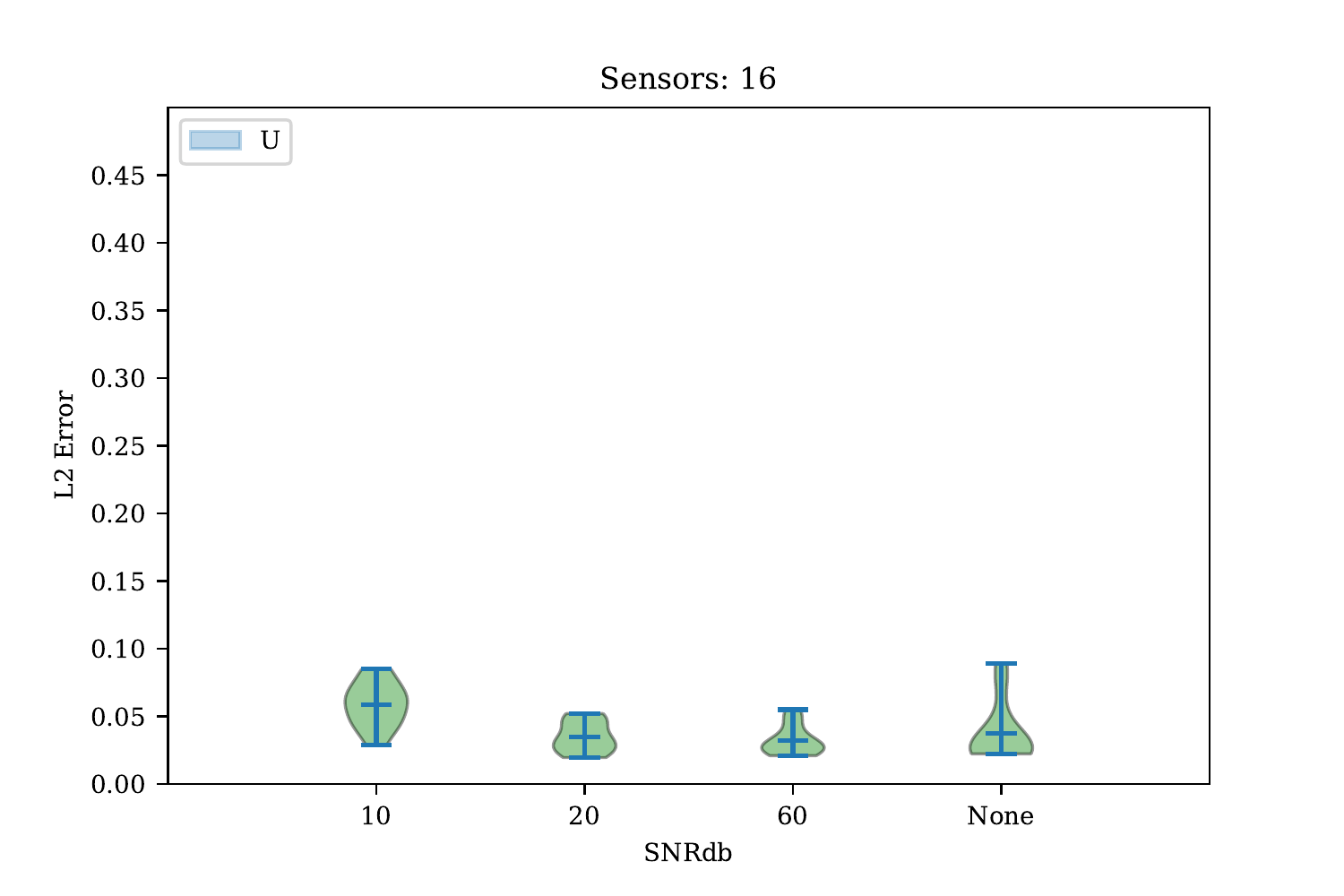}}
                \subfigure[]{ \includegraphics[width=0.45\textwidth]{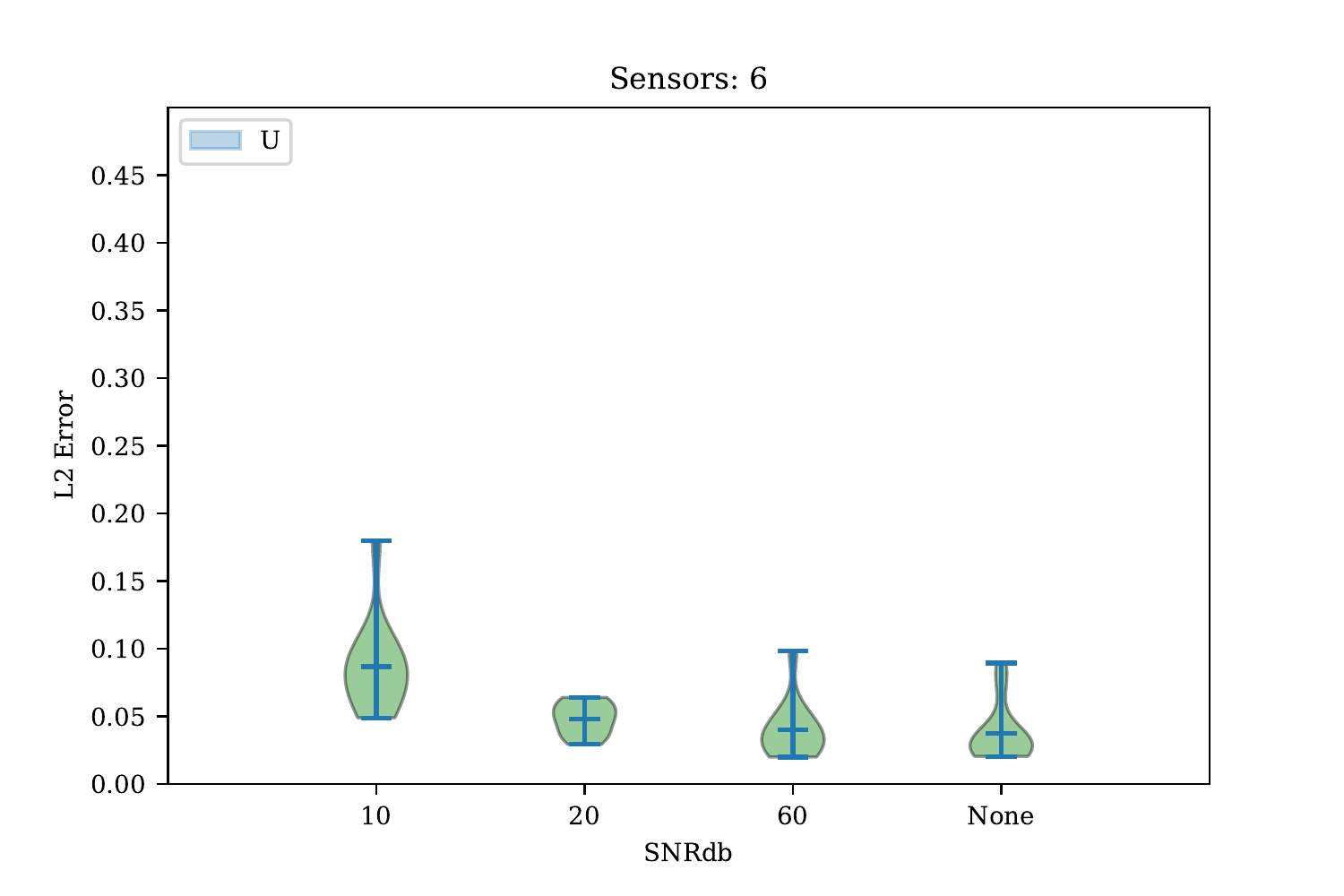}}
                
                \caption{Violin plot representing error $\epsilon$ distribution in prediction vector of data, for different noise levels in sensor measurements for Allen–Cahn equation test case.}
                \label{fig:AC_L2}
            \end{figure}

            Fig. \ref{fig:AC_plot} shows the prediction of Ensers with 16 sensors. A noticeable benefit of the continuous formulation is that prediction is smooth compared to discrete cases. Fig. \ref{fig:AC_L2} shows violin plot of L2 error defined in Eq. (\ref{eq:error}) between target and prediction with various noise levels in sensor measurement for the test case. In Eq. (\ref{eq:error}), $m=0$ and $\gamma^*=2$. Similar to discrete cases, the model is robust to noise in measurements. Fig. \ref{fig:AC_L2} shows a marginal increase in error with noise in the case of 16 sensors. Also, error bars with 6 sensors are similar to 16 sensors with low noise and increase slowly with noise.
            
    \subsection{Experiment: Convection-diffusion equations}\label{sec:CD}
        
        Next we consider a 2-dimensional linear variable-coefficient convection-diffusion equation on $\omega = [0, 2\pi]\times[0, 2\pi]$,
        
        \begin{equation}
            \label{eq:CDeq}
            u_t = a(x, y)u_x + b(x, y)u_y + cu_{xx} + du_{yy} \ (t, x, y) \in [0, 0.2] \times \omega 
        \end{equation}

        \begin{equation}
            \label{eq:CDcoeff}
            \begin{aligned}
            a(x, y) = 0.5( cos(y) + x(2\pi -x)sin(x) ) + 0.6 \\ 
            b(x, y) = 2( cos(y) + sin(x)  ) + 0.8 \\
            c = 0.2, d = 0.3
            \end{aligned}
        \end{equation}    
        
        Convection-diffusion equations are classical PDEs that are used to describe physical phenomena where particles, energy, or other physical quantities are transferred inside a physical system due to two processes namely diffusion and convection. These equations are widely applied in many scientific areas and industrial fields, such as pollutants dispersion in rivers or atmosphere, solute transferring in a porous medium, and oil reservoir simulation. We consider variables convection and diffusion coefficients Eq. (\ref{eq:CDcoeff}) in this experiment.  
            
        \subsubsection{Data-set and Model Parameters}
        
\begin{figure}[!htb]
                \centering
                \includegraphics[width=1\linewidth]{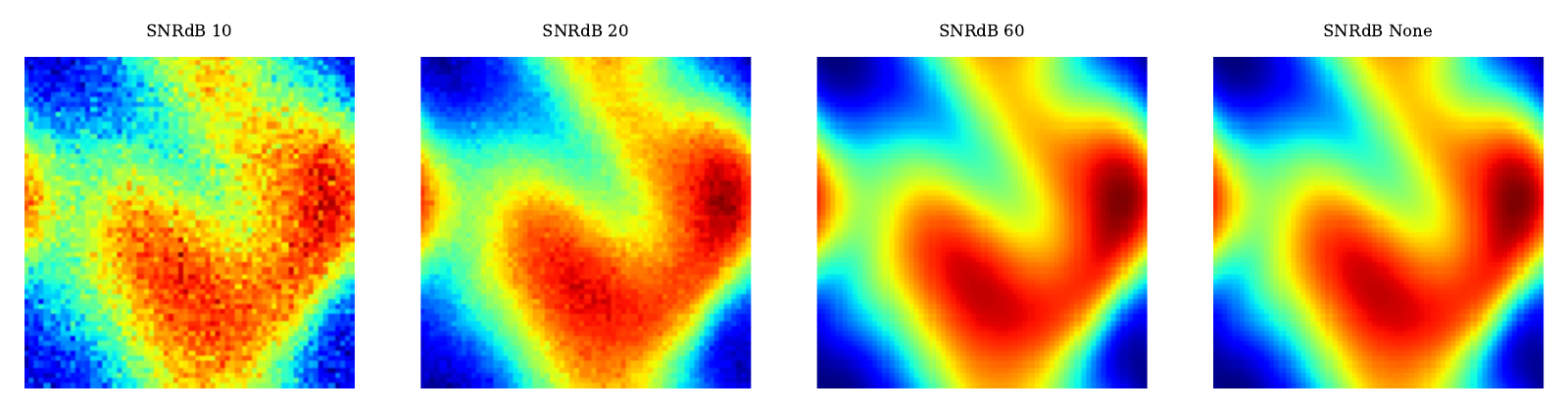}
                \caption{Data plots of Convection-diffusion equation for visualizing different noise levels. SNRdB represents signal to noise ratio in decibels.}
                \label{fig:CDNoise}
            \end{figure}

            Data is generated by solving the problem (\ref{eq:CDeq}) using a high precision numerical scheme with a pseudo-spectral method for spatial discretization and 4th order Runge-Kutta for temporal discretization (with time step size $\delta t = 0.01$). We assume periodic boundary conditions and the initial value
            
            \begin{equation}
                \label{eq:CDinitCond}
                u(x, y, 0) = \sum_{|k|, |l| \leq N} \lambda_{k,l} cos(kx + ly) + \gamma_{k, l} sin(kx + ly)
            \end{equation}
            
            where $N=9, \lambda_{k,l}, \gamma_{k, l} \sim \mathcal{N} (0, 0.02)$, and k and l are chosen randomly. For more details on the data-set see \cite{long18a}. Noisy data used for sensor measurement during testing is shown in Fig. \ref{fig:CDNoise} with various noise levels at a particular time. We use a simulation of length 40 as shown in table \ref{tab:CDDataParam} along with other data parameters.
            
            \begin{table}[!htb]
                \centering
                \caption{Data-set parameters for Convection-diffusion equation.}
                \label{tab:CDDataParam}
                
                \begin{tabular}{|p{1cm}|p{1cm}|p{1cm}|p{1cm}|p{1cm}|p{2cm}|}
                    \hline
                    
                    $\gamma$& $M$   & $L$   & $z$   & $\omega$  & domain        \\ \hline
                    5       & 1     & 40    & 1     & 4096      & $64\times64$  \\ \hline
                \end{tabular}
            \end{table}

            \begin{table}[!htb]
                \centering
                \caption{Network architecture of proposed Ensers for Convection-diffusion equation.}
                \label{tab:CDnet}
                
                \begin{tabular}{|p{2cm}|p{2cm}|p{2cm}|p{2cm}|}
                    \hline
                    
                    Layer   & Input & Output            & Activation    \\ \hline
                    FC      & 8    & 128                & Tanh          \\ \hline
                    FC      & 128  & 256                & Tanh          \\ \hline
                    FC      & 256  & 256                & Tanh          \\ \hline
                    FC      & 256  & 128                & Tanh          \\ \hline
                    FC      & 128  & $M*\gamma$         & linear        \\ \hline
                    
                \end{tabular}
            \end{table}
        
            \begin{table}[!htb]
                \centering
                \caption{Training hyper parameters of Ensers for Convection-diffusion equation.}
                \label{tab:CDTrainHP}
                
                \begin{tabular}{|p{0.9cm}|p{0.9cm}|p{0.9cm}|p{0.9cm}|p{0.9cm}|p{0.9cm}|p{0.9cm}|p{0.9cm}|p{0.9cm}|p{0.9cm}|p{0.9cm}|p{0.9cm}|}
                    \hline
                    
                    $\eta_o$& $\eta_{i0}$   & $\hat{\eta_{i}}$  & $\beta$   & $I_o$ & $I_i$ & $\zeta_0$ & $\hat{\zeta}$     & $p$   & $h$   & $N$ & $\varsigma$   \\ \hline
                    0.0005  & 0.005         & $5\mathrm{e}{-5}$ & 11        & 1001  & 10    & 0.005     & $1\mathrm{e}{-5}$ & 32    & 1024   & 33  & 6             \\ \hline
                \end{tabular}
            \end{table}
            
            \begin{table}[!htb]
                \centering
                \caption{Testing hyper parameters of Ensers for Convection-diffusion equation. $\eta_i$: inner learning rate, $I_i$: inner iterations}
                \label{tab:CDTestHP}
                
                \begin{tabular}{|p{1cm}|p{1cm}|p{1.5cm}|p{1cm}|}
                    \hline
                    
                    $\eta_i$  & $I_i$   & $p$       & $\hat{N}$ \\ \hline
                    0.2       & 50      & [4, 16]   & 12         \\ \hline
                \end{tabular}
            \end{table}
            
            Network architecture is shown in table \ref{tab:CDnet}. Similar to the previous case a deep network with $5$ layers and a small output size is used. The activation function is Tanh, which also has a continuous first derivative. For training, we use $p=1024$ training nodes i.e. $25\%$ of total nodes. Training and testing hyperparameters are shown in tables k and ll respectively. Inner loop learning rate $\eta_i$ and physics loss penalty $\zeta$ are increased linearly with each epoch.

        \subsubsection{Results and Discussions}
            
            \begin{figure}[!htb]
                \centering
                \includegraphics[width=1\linewidth]{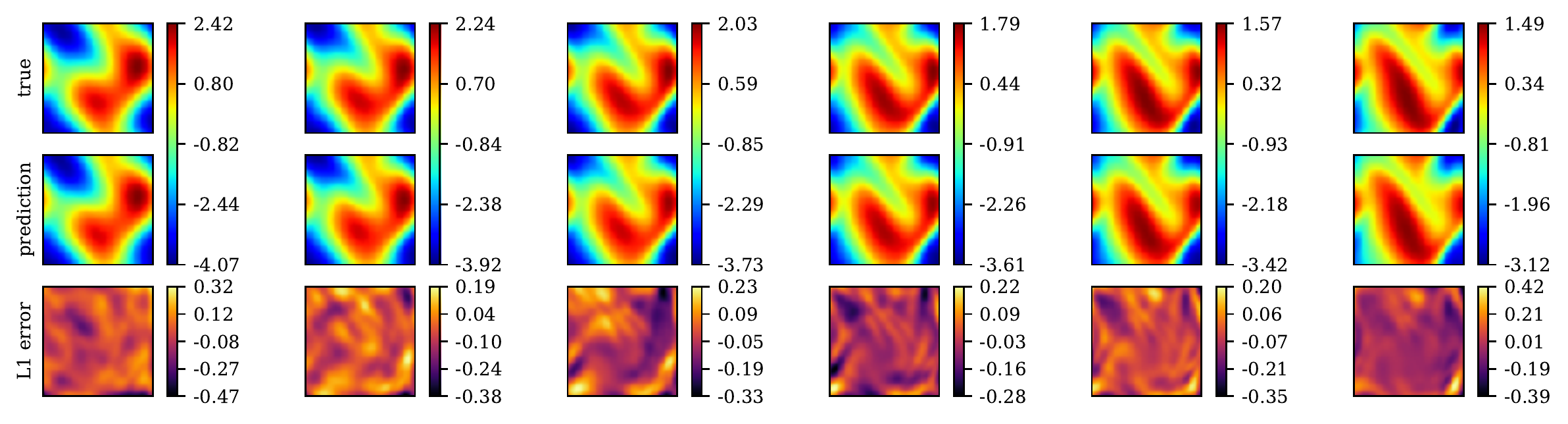}
                \caption{Ensers predictions of a Convection-diffusion equation test case. (Top to bottom) Target solution, Ensers prediction, L1error.}
                \label{fig:CD_plot}
            \end{figure}
            
            \begin{figure}
                \centering
                \subfigure[]{ \includegraphics[width=0.45\textwidth]{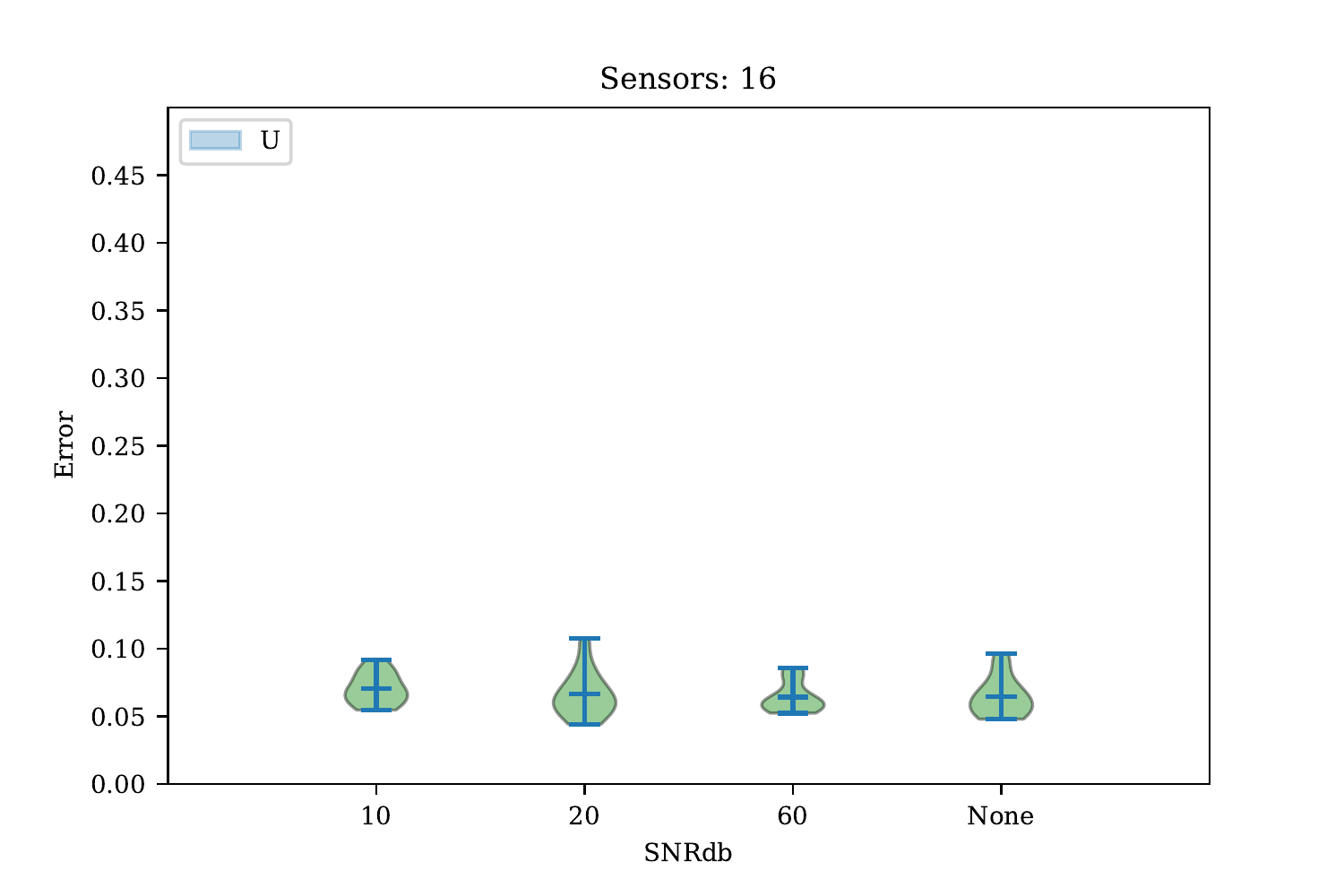}}
                \subfigure[]{ \includegraphics[width=0.45\textwidth]{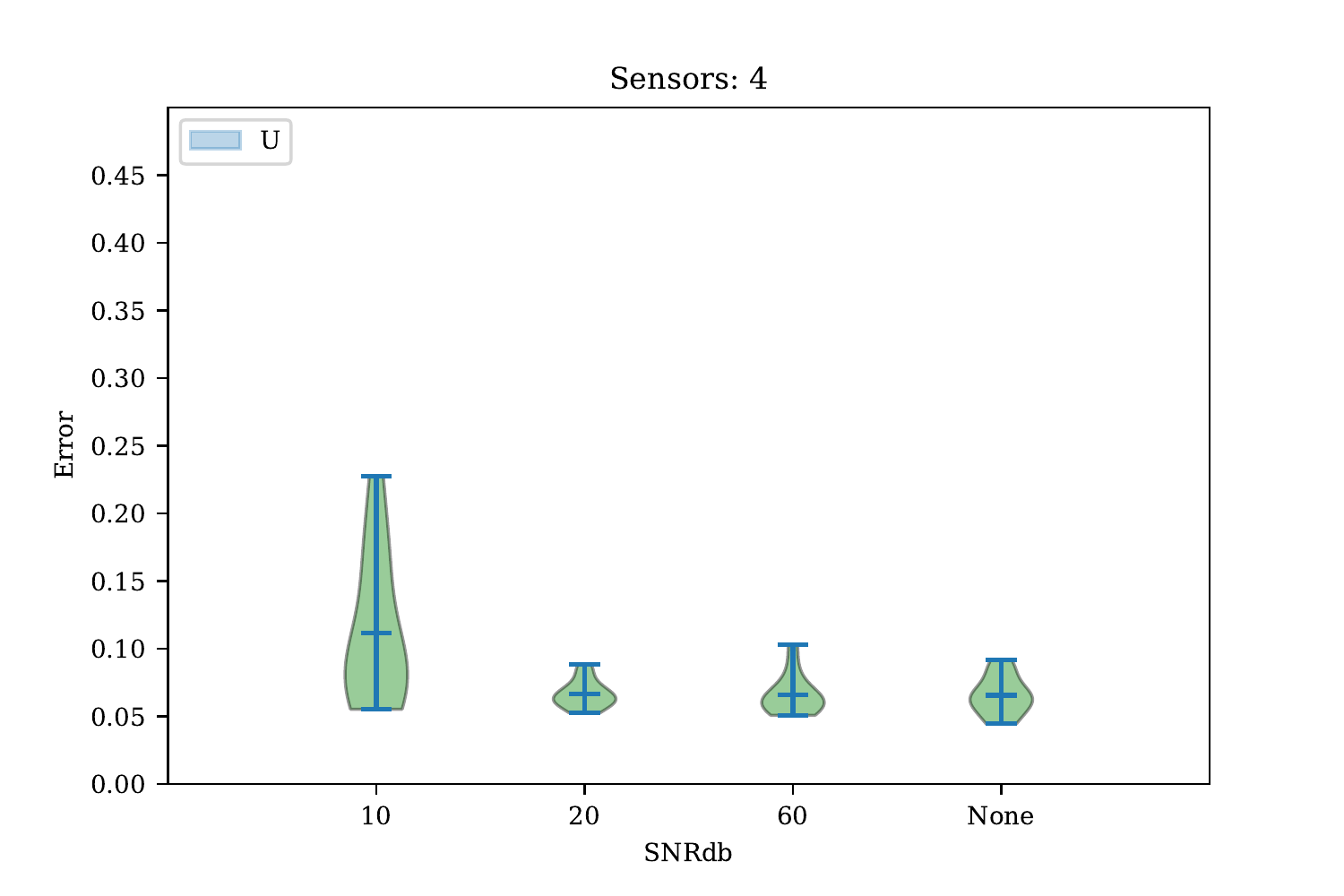}}
                
                \caption{Violin plot representing error $\epsilon$ distribution in prediction vector of x,y velocity and pressure for different noise levels in sensor measurements for Convection-diffusion equation test case.}
                \label{fig:CD_L2}
            \end{figure}
            
            Fig. \ref{fig:CD_plot} shows the prediction of the Ensers with 16 sensors. The model is able to predict the state accurately and with little distortion. Fig. \ref{fig:CD_L2} shows violin plot of L2 error defined in Eq. (\ref{eq:error}) between target and prediction with various noise levels in sensor measurement for the test case. In Eq. (\ref{eq:error}), $m=0$ and $\gamma^*=2$ i.e. we use middle prediction out $\gamma$ time steps. We see the model is robust to noise as the mean error in the plots are close for cases of no noise($SNR_{dB}=None$) and high noise($SNR_{dB}=10$) for 16 sensors case and increases slightly noise for 4 sensor cases. Training continuous model for two dimensions takes more time than discrete because of the increased number of collocation points and multiple automatic differentiation required for calculating loss function.

\section{Conclusions}\label{sec:conc}
    In this work, we develop a novel technique to learn a deep learning model from spare moving training labels for which model reduction is nontrivial. The method uses an implicit optimization layer for minimizing the energy of the solution implemented through the technique of unrolled differentiation. We proposed two formulations based on this technique for discrete and continuous prediction in space. For learning from spare training labels we included a physics-based loss function calculated via convolutional filters in the discrete formulation and via automatic differentiation in the continuous formulation. Where most deep learning-based methods assume fixed sensors Ensers is capable of predicting full states given a varied number of sensors at random locations.
    
    We demonstrate the model performance using two-fluid problems of 2D coupled Burgers’ equation and Flow Past Cylinder for the discrete case. For the continuous case, we used two problems namely Allen–Cahn equation and the Convection-diffusion equation. Model is shown to be robust against noisy sensor measurements. Future work can be aimed at quantifying uncertainty in such networks. Another direction can be to train networks for future states only using initial and boundary conditions.

\section*{Acknowledgements}
    SC acknowledges the financial support received from Science and Engineering Research Board (SERB) via project no. SRG/2021/000467 and Indian Institute Of Technology–Delhi in form of seed grant.
    
\section*{Reproducibility}
    The codes associated with the paper will be released on acceptance. 

\appendix
    \section{Convolution operators for gradient and laplacian terms}
    \label{appA}
    Sobel Filter used to estimate 1st-order gradient is:
    \begin{subequations}
        \label{eq:gradientConv}
        \begin{equation}
            E = \left[ \begin{array}{ccccc}
                    1   & -8    & 0     & 8     & -1  \\
                    2   & -16   & 0     & 16    & -2  \\
                    3   & -24   & 0     & 24    & -3  \\
                    2   & -16   & 0     & 16    & -2  \\
                    1   & -8    & 0     & 8     & -1  \\
                \end{array} \right]
        \end{equation}
        
        \begin{equation}
            \frac{\partial}{\partial x}  = E \times \frac{1}{9*12\delta x}
        \end{equation}
        
        Filter used to estimate laplacian is:
        \begin{equation}
            \frac{\partial}{\partial y} = E^T \times \frac{1}{9*12\delta y}
        \end{equation}
    \end{subequations}
    
    \begin{equation}
        \label{eq:gradientConvX}
        \frac{\partial^2}{\partial x^2} + \frac{\partial^2}{\partial y^2} filter = 
        \left[ \begin{array}{ccccc}
            0   & 0     & -1    & 0     & 0  \\
            0   & 0     & 16    & 0     & 0  \\
            -1  & 16    & -60   & 16    & -1 \\
            0   & 0     & 16    & 0     & 0  \\
            0   & 0     & -1    & 0     & 0  \\
        \end{array} \right] \times \frac{1}{12\delta x \delta y}
    \end{equation}

\bibliographystyle{elsarticle-num-names}
\bibliography{sample}

\begin{thebibliography}{50}
\expandafter\ifx\csname natexlab\endcsname\relax\def\natexlab#1{#1}\fi
\providecommand{\url}[1]{\texttt{#1}}
\providecommand{\href}[2]{#2}
\providecommand{\path}[1]{#1}
\providecommand{\DOIprefix}{doi:}
\providecommand{\ArXivprefix}{arXiv:}
\providecommand{\URLprefix}{URL: }
\providecommand{\Pubmedprefix}{pmid:}
\providecommand{\doi}[1]{\href{http://dx.doi.org/#1}{\path{#1}}}
\providecommand{\Pubmed}[1]{\href{pmid:#1}{\path{#1}}}
\providecommand{\bibinfo}[2]{#2}
\ifx\xfnm\relax \def\xfnm[#1]{\unskip,\space#1}\fi
\bibitem[{Cordier(2011)}]{Cordier2011}
\bibinfo{author}{L.~Cordier}, \bibinfo{publisher}{Springer Vienna},
  \bibinfo{year}{2011}. \DOIprefix\doi{10.1007/978-3-7091-0758-4_1}.
\bibitem[{Semaan et~al.(2019)Semaan, El~Sayed, and Radespiel}]{Semaan}
\bibinfo{author}{R.~Semaan}, \bibinfo{author}{M.~Y. El~Sayed},
  \bibinfo{author}{R.~Radespiel}, \bibinfo{publisher}{Springer International
  Publishing}, \bibinfo{address}{Cham}, \bibinfo{year}{2019}.
\bibitem[{Sankaran et~al.(2012)Sankaran, Esmaily~Moghadam, Kahn, Tseng,
  Guccione, and Marsden}]{pmid22539149}
\bibinfo{author}{S.~Sankaran}, \bibinfo{author}{M.~Esmaily~Moghadam},
  \bibinfo{author}{A.~M. Kahn}, \bibinfo{author}{E.~E. Tseng},
  \bibinfo{author}{J.~M. Guccione}, \bibinfo{author}{A.~L. Marsden},
\newblock \bibinfo{title}{{{P}atient-specific multiscale modeling of blood flow
  for coronary artery bypass graft surgery}},
\newblock \bibinfo{journal}{Ann Biomed Eng} \bibinfo{volume}{40}
  (\bibinfo{year}{2012}) \bibinfo{pages}{2228--2242}.
\bibitem[{Yakhot et~al.(2007)Yakhot, Anor, and Karniadakis}]{pmid18092738}
\bibinfo{author}{A.~Yakhot}, \bibinfo{author}{T.~Anor}, \bibinfo{author}{G.~E.
  Karniadakis},
\newblock \bibinfo{title}{{{A} reconstruction method for gappy and noisy
  arterial flow data}},
\newblock \bibinfo{journal}{IEEE Trans Med Imaging} \bibinfo{volume}{26}
  (\bibinfo{year}{2007}) \bibinfo{pages}{1681--1697}.
\bibitem[{Kissas et~al.(2020)Kissas, Yang, Hwuang, Witschey, Detre, and
  Perdikaris}]{KISSAS2}
\bibinfo{author}{G.~Kissas}, \bibinfo{author}{Y.~Yang},
  \bibinfo{author}{E.~Hwuang}, \bibinfo{author}{W.~R. Witschey},
  \bibinfo{author}{J.~A. Detre}, \bibinfo{author}{P.~Perdikaris},
\newblock \bibinfo{title}{Machine learning in cardiovascular flows modeling:
  Predicting arterial blood pressure from non-invasive 4d flow mri data using
  physics-informed neural networks},
\newblock \bibinfo{journal}{Computer Methods in Applied Mechanics and
  Engineering} \bibinfo{volume}{358} (\bibinfo{year}{2020})
  \bibinfo{pages}{112623}. \URLprefix
  \url{https://www.sciencedirect.com/science/article/pii/S0045782519305055}.
  \DOIprefix\doi{https://doi.org/10.1016/j.cma.2019.112623}.
\bibitem[{Graziano et~al.(2016)Graziano, D’Errico, and
  Rufino}]{GRAZIANO201672}
\bibinfo{author}{M.~D. Graziano}, \bibinfo{author}{M.~D’Errico},
  \bibinfo{author}{G.~Rufino},
\newblock \bibinfo{title}{Ship heading and velocity analysis by wake detection
  in sar images},
\newblock \bibinfo{journal}{Acta Astronautica} \bibinfo{volume}{128}
  (\bibinfo{year}{2016}) \bibinfo{pages}{72--82}. \URLprefix
  \url{https://www.sciencedirect.com/science/article/pii/S0094576516300303}.
  \DOIprefix\doi{https://doi.org/10.1016/j.actaastro.2016.07.001}.
\bibitem[{Kalnay(2002)}]{kalnay_2002}
\bibinfo{author}{E.~Kalnay}, \bibinfo{title}{Atmospheric Modeling, Data
  Assimilation and Predictability}, \bibinfo{publisher}{Cambridge University
  Press}, \bibinfo{year}{2002}. \DOIprefix\doi{10.1017/CBO9780511802270}.
\bibitem[{Ong et~al.(2015)Ong, Uecker, Tariq, Hsiao, Alley, Vasanawala, and
  Lustig}]{Ong}
\bibinfo{author}{F.~Ong}, \bibinfo{author}{M.~Uecker},
  \bibinfo{author}{U.~Tariq}, \bibinfo{author}{A.~Hsiao},
  \bibinfo{author}{M.~T. Alley}, \bibinfo{author}{S.~S. Vasanawala},
  \bibinfo{author}{M.~Lustig},
\newblock \bibinfo{title}{Robust 4d flow denoising using divergence-free
  wavelet transform},
\newblock \bibinfo{journal}{Magnetic Resonance in Medicine}
  \bibinfo{volume}{73} (\bibinfo{year}{2015}) \bibinfo{pages}{828--842}.
  \URLprefix \url{https://onlinelibrary.wiley.com/doi/abs/10.1002/mrm.25176}.
  \DOIprefix\doi{https://doi.org/10.1002/mrm.25176}.
  \href{http://arxiv.org/abs/https://onlinelibrary.wiley.com/doi/pdf/10.1002/mrm.25176}{{\tt
  arXiv:https://onlinelibrary.wiley.com/doi/pdf/10.1002/mrm.25176}}.
\bibitem[{Callaghan and Grieve(2017)}]{pmid27885707}
\bibinfo{author}{F.~M. Callaghan}, \bibinfo{author}{S.~M. Grieve},
\newblock \bibinfo{title}{{{S}patial resolution and velocity field improvement
  of 4{D}-flow {M}{R}{I}}},
\newblock \bibinfo{journal}{Magn Reson Med} \bibinfo{volume}{78}
  (\bibinfo{year}{2017}) \bibinfo{pages}{1959--1968}.
\bibitem[{Fathi et~al.(2018)Fathi, Bakhshinejad, Baghaie, Saloner, Sacho, Rayz,
  and D'Souza}]{pmid30423501}
\bibinfo{author}{M.~F. Fathi}, \bibinfo{author}{A.~Bakhshinejad},
  \bibinfo{author}{A.~Baghaie}, \bibinfo{author}{D.~Saloner},
  \bibinfo{author}{R.~H. Sacho}, \bibinfo{author}{V.~L. Rayz},
  \bibinfo{author}{R.~M. D'Souza},
\newblock \bibinfo{title}{{{D}enoising and spatial resolution enhancement of
  4{D} flow {M}{R}{I} using proper orthogonal decomposition and lasso
  regularization}},
\newblock \bibinfo{journal}{Comput Med Imaging Graph} \bibinfo{volume}{70}
  (\bibinfo{year}{2018}) \bibinfo{pages}{165--172}.
\bibitem[{Bishop(2006)}]{bishop2006pattern}
\bibinfo{author}{C.~M. Bishop}, \bibinfo{title}{Pattern recognition and machine
  learning}, \bibinfo{publisher}{springer}, \bibinfo{year}{2006}.
\bibitem[{Reif et~al.(1999)Reif, Gunther, Yaz, and
  Unbehauen}]{reif1999stochastic}
\bibinfo{author}{K.~Reif}, \bibinfo{author}{S.~Gunther},
  \bibinfo{author}{E.~Yaz}, \bibinfo{author}{R.~Unbehauen},
\newblock \bibinfo{title}{Stochastic stability of the discrete-time extended
  kalman filter},
\newblock \bibinfo{journal}{IEEE Transactions on Automatic control}
  \bibinfo{volume}{44} (\bibinfo{year}{1999}) \bibinfo{pages}{714--728}.
\bibitem[{Wan et~al.(2001)Wan, Van Der~Merwe, and Haykin}]{wan2001unscented}
\bibinfo{author}{E.~A. Wan}, \bibinfo{author}{R.~Van Der~Merwe},
  \bibinfo{author}{S.~Haykin},
\newblock \bibinfo{title}{The unscented kalman filter},
\newblock \bibinfo{journal}{Kalman filtering and neural networks}
  \bibinfo{volume}{5} (\bibinfo{year}{2001}) \bibinfo{pages}{221--280}.
\bibitem[{Tu et~al.(2012)Tu, Griffin, Hart, Rowley, Cattafesta, and
  Ukeiley}]{Jonathan}
\bibinfo{author}{J.~Tu}, \bibinfo{author}{J.~Griffin},
  \bibinfo{author}{A.~Hart}, \bibinfo{author}{C.~Rowley},
  \bibinfo{author}{L.~Cattafesta}, \bibinfo{author}{L.~Ukeiley},
\newblock \bibinfo{title}{Integration of non-time-resolved piv and
  time-resolved velocity point sensors for dynamic estimation of velocity
  fields},
\newblock \bibinfo{journal}{Experiments in Fluids} \bibinfo{volume}{54}
  (\bibinfo{year}{2012}). \DOIprefix\doi{10.1007/s00348-012-1429-7}.
\bibitem[{{Schmid} and {Sesterhenn}(2008)}]{2008AP}
\bibinfo{author}{P.~{Schmid}}, \bibinfo{author}{J.~{Sesterhenn}},
\newblock \bibinfo{title}{{Dynamic Mode Decomposition of numerical and
  experimental data}},
\newblock in: \bibinfo{booktitle}{APS Division of Fluid Dynamics Meeting
  Abstracts}, volume~\bibinfo{volume}{61} of \textit{\bibinfo{series}{APS
  Meeting Abstracts}}, \bibinfo{year}{2008}, p. \bibinfo{pages}{MR.007}.
\bibitem[{ROWLEY et~al.(2009)ROWLEY, MEZIĆ, BAGHERI, SCHLATTER, and
  HENNINGSON}]{rowley_mez}
\bibinfo{author}{C.~W. ROWLEY}, \bibinfo{author}{I.~MEZIĆ},
  \bibinfo{author}{S.~BAGHERI}, \bibinfo{author}{P.~SCHLATTER},
  \bibinfo{author}{D.~S. HENNINGSON},
\newblock \bibinfo{title}{Spectral analysis of nonlinear flows},
\newblock \bibinfo{journal}{Journal of Fluid Mechanics} \bibinfo{volume}{641}
  (\bibinfo{year}{2009}) \bibinfo{pages}{115–127}.
  \DOIprefix\doi{10.1017/S0022112009992059}.
\bibitem[{Buffoni et~al.(2008)Buffoni, Camarri, Iollo, Lombardi, and
  Salvetti}]{BUFFONI20082626}
\bibinfo{author}{M.~Buffoni}, \bibinfo{author}{S.~Camarri},
  \bibinfo{author}{A.~Iollo}, \bibinfo{author}{E.~Lombardi},
  \bibinfo{author}{M.~Salvetti},
\newblock \bibinfo{title}{A non-linear observer for unsteady three-dimensional
  flows},
\newblock \bibinfo{journal}{Journal of Computational Physics}
  \bibinfo{volume}{227} (\bibinfo{year}{2008}) \bibinfo{pages}{2626 -- 2643}.
  \URLprefix
  \url{http://www.sciencedirect.com/science/article/pii/S0021999107004858}.
  \DOIprefix\doi{https://doi.org/10.1016/j.jcp.2007.11.005}.
\bibitem[{Adrian(1975)}]{Adrian}
\bibinfo{author}{R.~Adrian}, \bibinfo{title}{On the role of conditional
  averages in turbulence theory.}, \bibinfo{year}{1975}.
\bibitem[{Guezennec(1989)}]{Guezennec}
\bibinfo{author}{Y.~G. Guezennec},
\newblock \bibinfo{title}{Stochastic estimation of coherent structures in
  turbulent boundary layers},
\newblock \bibinfo{journal}{Physics of Fluids A: Fluid Dynamics}
  \bibinfo{volume}{1} (\bibinfo{year}{1989}) \bibinfo{pages}{1054--1060}.
  \URLprefix \url{https://doi.org/10.1063/1.857396}.
  \DOIprefix\doi{10.1063/1.857396}.
  \href{http://arxiv.org/abs/https://doi.org/10.1063/1.857396}{{\tt
  arXiv:https://doi.org/10.1063/1.857396}}.
\bibitem[{Ewing and Citriniti(1999)}]{Ewing}
\bibinfo{author}{D.~Ewing}, \bibinfo{author}{J.~H. Citriniti},
\newblock \bibinfo{title}{Examination of a lse/pod complementary technique
  using single and multi-time information in the axisymmetric shear layer},
\newblock in: \bibinfo{editor}{J.~N. S{\o}rensen}, \bibinfo{editor}{E.~J.
  Hopfinger}, \bibinfo{editor}{N.~Aubry} (Eds.), \bibinfo{booktitle}{IUTAM
  Symposium on Simulation and Identification of Organized Structures in Flows},
  \bibinfo{publisher}{Springer Netherlands}, \bibinfo{address}{Dordrecht},
  \bibinfo{year}{1999}, pp. \bibinfo{pages}{375--384}.
\bibitem[{Naguib et~al.(2001)Naguib, Wark, and Juckenhöfel}]{Naguib}
\bibinfo{author}{A.~M. Naguib}, \bibinfo{author}{C.~E. Wark},
  \bibinfo{author}{O.~Juckenhöfel},
\newblock \bibinfo{title}{Stochastic estimation and flow sources associated
  with surface pressure events in a turbulent boundary layer},
\newblock \bibinfo{journal}{Physics of Fluids} \bibinfo{volume}{13}
  (\bibinfo{year}{2001}) \bibinfo{pages}{2611--2626}. \URLprefix
  \url{https://doi.org/10.1063/1.1389284}. \DOIprefix\doi{10.1063/1.1389284}.
  \href{http://arxiv.org/abs/https://doi.org/10.1063/1.1389284}{{\tt
  arXiv:https://doi.org/10.1063/1.1389284}}.
\bibitem[{Bonnet et~al.(1994)Bonnet, Cole, Delville, Glauser, and
  Ukeiley}]{Bonnet1994}
\bibinfo{author}{J.~P. Bonnet}, \bibinfo{author}{D.~R. Cole},
  \bibinfo{author}{J.~Delville}, \bibinfo{author}{M.~N. Glauser},
  \bibinfo{author}{L.~S. Ukeiley},
\newblock \bibinfo{title}{Stochastic estimation and proper orthogonal
  decomposition: Complementary techniques for identifying structure},
\newblock \bibinfo{journal}{Experiments in Fluids} \bibinfo{volume}{17}
  (\bibinfo{year}{1994}) \bibinfo{pages}{307--314}. \URLprefix
  \url{https://doi.org/10.1007/BF01874409}. \DOIprefix\doi{10.1007/BF01874409}.
\bibitem[{Pinier et~al.(2007)Pinier, Ausseur, Glauser, and Higuchi}]{Aero}
\bibinfo{author}{J.~Pinier}, \bibinfo{author}{J.~Ausseur},
  \bibinfo{author}{M.~Glauser}, \bibinfo{author}{H.~Higuchi},
\newblock \bibinfo{title}{Proportional closed-loop feedback control of flow
  separation},
\newblock \bibinfo{journal}{Aiaa Journal - AIAA J} \bibinfo{volume}{45}
  (\bibinfo{year}{2007}) \bibinfo{pages}{181--190}.
  \DOIprefix\doi{10.2514/1.23465}.
\bibitem[{Adrian(1979)}]{tur}
\bibinfo{author}{R.~Adrian},
\newblock \bibinfo{title}{Conditional eddies in isotropic turbulence},
\newblock \bibinfo{journal}{Physics of Fluids} \bibinfo{volume}{22}
  (\bibinfo{year}{1979}). \DOIprefix\doi{10.1063/1.862515}.
\bibitem[{Tung and Adrian(1980)}]{Tung}
\bibinfo{author}{T.~C. Tung}, \bibinfo{author}{R.~J. Adrian},
\newblock \bibinfo{title}{Higher‐order estimates of conditional eddies in
  isotropic turbulence},
\newblock \bibinfo{journal}{The Physics of Fluids} \bibinfo{volume}{23}
  (\bibinfo{year}{1980}) \bibinfo{pages}{1469--1470}. \URLprefix
  \url{https://aip.scitation.org/doi/abs/10.1063/1.863130}.
  \DOIprefix\doi{10.1063/1.863130}.
  \href{http://arxiv.org/abs/https://aip.scitation.org/doi/pdf/10.1063/1.863130}{{\tt
  arXiv:https://aip.scitation.org/doi/pdf/10.1063/1.863130}}.
\bibitem[{Kumar et~al.(2021)Kumar, Bahl, and Chakraborty}]{kumar2021state}
\bibinfo{author}{Y.~Kumar}, \bibinfo{author}{P.~Bahl},
  \bibinfo{author}{S.~Chakraborty}, \bibinfo{title}{State estimation with
  limited sensors -- a deep learning based approach}, \bibinfo{year}{2021}.
  \href{http://arxiv.org/abs/2101.11513}{{\tt arXiv:2101.11513}}.
\bibitem[{Erichson et~al.(2019)Erichson, Mathelin, Yao, Brunton, Mahoney, and
  Kutz}]{erichson2019shallow}
\bibinfo{author}{N.~B. Erichson}, \bibinfo{author}{L.~Mathelin},
  \bibinfo{author}{Z.~Yao}, \bibinfo{author}{S.~L. Brunton},
  \bibinfo{author}{M.~W. Mahoney}, \bibinfo{author}{J.~N. Kutz},
  \bibinfo{title}{Shallow learning for fluid flow reconstruction with limited
  sensors and limited data}, \bibinfo{year}{2019}.
  \href{http://arxiv.org/abs/1902.07358}{{\tt arXiv:1902.07358}}.
\bibitem[{Nair and Goza(2020)}]{nair_goza_2020}
\bibinfo{author}{N.~J. Nair}, \bibinfo{author}{A.~Goza},
\newblock \bibinfo{title}{Leveraging reduced-order models for state estimation
  using deep learning},
\newblock \bibinfo{journal}{Journal of Fluid Mechanics} \bibinfo{volume}{897}
  (\bibinfo{year}{2020}) \bibinfo{pages}{R1}.
  \DOIprefix\doi{10.1017/jfm.2020.409}.
\bibitem[{Gao et~al.(2021)Gao, Sun, and Wang}]{gaoHan}
\bibinfo{author}{H.~Gao}, \bibinfo{author}{L.~Sun}, \bibinfo{author}{J.-X.
  Wang},
\newblock \bibinfo{title}{Super-resolution and denoising of fluid flow using
  physics-informed convolutional neural networks without high-resolution
  labels},
\newblock \bibinfo{journal}{Physics of Fluids} \bibinfo{volume}{33}
  (\bibinfo{year}{2021}) \bibinfo{pages}{073603}. \URLprefix
  \url{https://doi.org/10.1063/5.0054312}. \DOIprefix\doi{10.1063/5.0054312}.
  \href{http://arxiv.org/abs/https://doi.org/10.1063/5.0054312}{{\tt
  arXiv:https://doi.org/10.1063/5.0054312}}.
\bibitem[{Grefenstette et~al.(2019)Grefenstette, Amos, Yarats, Htut, Molchanov,
  Meier, Kiela, Cho, and Chintala}]{grefen}
\bibinfo{author}{E.~Grefenstette}, \bibinfo{author}{B.~Amos},
  \bibinfo{author}{D.~Yarats}, \bibinfo{author}{P.~M. Htut},
  \bibinfo{author}{A.~Molchanov}, \bibinfo{author}{F.~Meier},
  \bibinfo{author}{D.~Kiela}, \bibinfo{author}{K.~Cho},
  \bibinfo{author}{S.~Chintala},
\newblock \bibinfo{title}{Generalized inner loop meta-learning},
\newblock \bibinfo{journal}{arXiv preprint arXiv:1910.01727}
  (\bibinfo{year}{2019}).
\bibitem[{Amos(2019)}]{amos2019}
\bibinfo{author}{B.~Amos}, \bibinfo{title}{{Differentiable Optimization-Based
  Modeling for Machine Learning}}, Ph.D. thesis, Carnegie Mellon University,
  \bibinfo{year}{2019}.
\bibitem[{Goodfellow et~al.(2013)Goodfellow, Mirza, Courville, and
  Bengio}]{Multipred}
\bibinfo{author}{I.~Goodfellow}, \bibinfo{author}{M.~Mirza},
  \bibinfo{author}{A.~Courville}, \bibinfo{author}{Y.~Bengio},
\newblock \bibinfo{title}{Multi-prediction deep boltzmann machines},
\newblock \bibinfo{journal}{Advances in Neural Information Processing Systems}
  (\bibinfo{year}{2013}).
\bibitem[{Stoyanov et~al.(2011)Stoyanov, Ropson, and Eisner}]{stoyanov11a}
\bibinfo{author}{V.~Stoyanov}, \bibinfo{author}{A.~Ropson},
  \bibinfo{author}{J.~Eisner},
\newblock \bibinfo{title}{Empirical risk minimization of graphical model
  parameters given approximate inference, decoding, and model structure},
\newblock in: \bibinfo{editor}{G.~Gordon}, \bibinfo{editor}{D.~Dunson},
  \bibinfo{editor}{M.~Dudík} (Eds.), \bibinfo{booktitle}{Proceedings of the
  Fourteenth International Conference on Artificial Intelligence and
  Statistics}, volume~\bibinfo{volume}{15} of
  \textit{\bibinfo{series}{Proceedings of Machine Learning Research}},
  \bibinfo{publisher}{PMLR}, \bibinfo{address}{Fort Lauderdale, FL, USA},
  \bibinfo{year}{2011}, pp. \bibinfo{pages}{725--733}. \URLprefix
  \url{https://proceedings.mlr.press/v15/stoyanov11a.html}.
\bibitem[{Brakel et~al.(2013)Brakel, Stroob, t, and Schrauwen}]{brakel13a}
\bibinfo{author}{P.~Brakel}, \bibinfo{author}{D.~Stroob}, \bibinfo{author}{t},
  \bibinfo{author}{B.~Schrauwen},
\newblock \bibinfo{title}{Training energy-based models for time-series
  imputation},
\newblock \bibinfo{journal}{Journal of Machine Learning Research}
  \bibinfo{volume}{14} (\bibinfo{year}{2013}) \bibinfo{pages}{2771--2797}.
  \URLprefix \url{http://jmlr.org/papers/v14/brakel13a.html}.
\bibitem[{Lecun et~al.(2006)Lecun, Chopra, and Hadsell}]{Lecun}
\bibinfo{author}{Y.~Lecun}, \bibinfo{author}{S.~Chopra},
  \bibinfo{author}{R.~Hadsell}, \bibinfo{title}{A tutorial on energy-based
  learning}, \bibinfo{year}{2006}.
\bibitem[{Utama et~al.(2018)Utama, N., and Iqbal}]{Utama}
\bibinfo{author}{D.~Utama}, \bibinfo{author}{A.~N.},
  \bibinfo{author}{M.~Iqbal},
\newblock \bibinfo{title}{An optimal generic model for multi-parameters and big
  data optimizing: a laboratory experimental study},
\newblock \bibinfo{journal}{Journal of Physics: Conference Series}
  \bibinfo{volume}{978} (\bibinfo{year}{2018}) \bibinfo{pages}{012045}.
  \DOIprefix\doi{10.1088/1742-6596/978/1/012045}.
\bibitem[{Belanger et~al.(2017)Belanger, Yang, and
  McCallum}]{belanger2017endtoend}
\bibinfo{author}{D.~Belanger}, \bibinfo{author}{B.~Yang},
  \bibinfo{author}{A.~McCallum}, \bibinfo{title}{End-to-end learning for
  structured prediction energy networks}, \bibinfo{year}{2017}.
  \href{http://arxiv.org/abs/1703.05667}{{\tt arXiv:1703.05667}}.
\bibitem[{Metz et~al.(2016)Metz, Poole, Pfau, and Sohl-Dickstein}]{Metz}
\bibinfo{author}{L.~Metz}, \bibinfo{author}{B.~Poole},
  \bibinfo{author}{D.~Pfau}, \bibinfo{author}{J.~Sohl-Dickstein},
\newblock \bibinfo{title}{Unrolled generative adversarial networks}
  (\bibinfo{year}{2016}).
\bibitem[{Johnson et~al.(2016)Johnson, Duvenaud, Wiltschko, Adams, and
  Datta}]{Johnson}
\bibinfo{author}{M.~J. Johnson}, \bibinfo{author}{D.~K. Duvenaud},
  \bibinfo{author}{A.~Wiltschko}, \bibinfo{author}{R.~P. Adams},
  \bibinfo{author}{S.~R. Datta},
\newblock \bibinfo{title}{Composing graphical models with neural networks for
  structured representations and fast inference},
\newblock in: \bibinfo{editor}{D.~Lee}, \bibinfo{editor}{M.~Sugiyama},
  \bibinfo{editor}{U.~Luxburg}, \bibinfo{editor}{I.~Guyon},
  \bibinfo{editor}{R.~Garnett} (Eds.), \bibinfo{booktitle}{Advances in Neural
  Information Processing Systems}, volume~\bibinfo{volume}{29},
  \bibinfo{publisher}{Curran Associates, Inc.}, \bibinfo{year}{2016}.
  \URLprefix
  \url{https://proceedings.neurips.cc/paper/2016/file/7d6044e95a16761171b130dcb476a43e-Paper.pdf}.
\bibitem[{Jordan-Squire(2015)}]{JordanSquire}
\bibinfo{author}{C.~Jordan-Squire},
\newblock \bibinfo{title}{Convex optimization over probability measures},
\newblock \bibinfo{year}{2015}.
\bibitem[{Raissi et~al.(2019)Raissi, Perdikaris, and Karniadakis}]{RAISSI_p}
\bibinfo{author}{M.~Raissi}, \bibinfo{author}{P.~Perdikaris},
  \bibinfo{author}{G.~Karniadakis},
\newblock \bibinfo{title}{Physics-informed neural networks: A deep learning
  framework for solving forward and inverse problems involving nonlinear
  partial differential equations},
\newblock \bibinfo{journal}{Journal of Computational Physics}
  \bibinfo{volume}{378} (\bibinfo{year}{2019}) \bibinfo{pages}{686--707}.
  \DOIprefix\doi{https://doi.org/10.1016/j.jcp.2018.10.045}.
\bibitem[{Zhu et~al.(2019)Zhu, Zabaras, Koutsourelakis, and
  Perdikaris}]{ZHU201956}
\bibinfo{author}{Y.~Zhu}, \bibinfo{author}{N.~Zabaras}, \bibinfo{author}{P.-S.
  Koutsourelakis}, \bibinfo{author}{P.~Perdikaris},
\newblock \bibinfo{title}{Physics-constrained deep learning for
  high-dimensional surrogate modeling and uncertainty quantification without
  labeled data},
\newblock \bibinfo{journal}{Journal of Computational Physics}
  \bibinfo{volume}{394} (\bibinfo{year}{2019}) \bibinfo{pages}{56--81}.
  \URLprefix
  \url{https://www.sciencedirect.com/science/article/pii/S0021999119303559}.
  \DOIprefix\doi{https://doi.org/10.1016/j.jcp.2019.05.024}.
\bibitem[{Geneva and Zabaras(2020)}]{GENEVA}
\bibinfo{author}{N.~Geneva}, \bibinfo{author}{N.~Zabaras},
\newblock \bibinfo{title}{Modeling the dynamics of pde systems with
  physics-constrained deep auto-regressive networks},
\newblock \bibinfo{journal}{Journal of Computational Physics}
  \bibinfo{volume}{403} (\bibinfo{year}{2020}) \bibinfo{pages}{109056}.
  \URLprefix
  \url{https://www.sciencedirect.com/science/article/pii/S0021999119307612}.
  \DOIprefix\doi{https://doi.org/10.1016/j.jcp.2019.109056}.
\bibitem[{Sobel and Feldman(1973)}]{Sobel}
\bibinfo{author}{I.~Sobel}, \bibinfo{author}{G.~Feldman},
\newblock \bibinfo{title}{A 3×3 isotropic gradient operator for image
  processing},
\newblock \bibinfo{journal}{Pattern Classification and Scene Analysis}
  (\bibinfo{year}{1973}) \bibinfo{pages}{271--272}.
\bibitem[{Logg et~al.(2012)Logg, Mardal, Wells et~al.}]{LoggMardalEtAl2012}
\bibinfo{author}{A.~Logg}, \bibinfo{author}{K.-A. Mardal},
  \bibinfo{author}{G.~N. Wells}, et~al., \bibinfo{title}{Automated Solution of
  Differential Equations by the Finite Element Method},
  \bibinfo{publisher}{Springer}, \bibinfo{year}{2012}.
  \DOIprefix\doi{10.1007/978-3-642-23099-8}.
\bibitem[{Glorot et~al.(2011)Glorot, Bordes, and Bengio}]{glorot11a}
\bibinfo{author}{X.~Glorot}, \bibinfo{author}{A.~Bordes},
  \bibinfo{author}{Y.~Bengio},
\newblock \bibinfo{title}{Deep sparse rectifier neural networks},
\newblock in: \bibinfo{editor}{G.~Gordon}, \bibinfo{editor}{D.~Dunson},
  \bibinfo{editor}{M.~Dudík} (Eds.), \bibinfo{booktitle}{Proceedings of the
  Fourteenth International Conference on Artificial Intelligence and
  Statistics}, volume~\bibinfo{volume}{15} of
  \textit{\bibinfo{series}{Proceedings of Machine Learning Research}},
  \bibinfo{publisher}{PMLR}, \bibinfo{address}{Fort Lauderdale, FL, USA},
  \bibinfo{year}{2011}, pp. \bibinfo{pages}{315--323}. \URLprefix
  \url{https://proceedings.mlr.press/v15/glorot11a.html}.
\bibitem[{Jasak(2009)}]{jasak2009openfoam}
\bibinfo{author}{H.~Jasak},
\newblock \bibinfo{title}{Openfoam: open source cfd in research and industry},
\newblock \bibinfo{journal}{International Journal of Naval Architecture and
  Ocean Engineering} \bibinfo{volume}{1} (\bibinfo{year}{2009})
  \bibinfo{pages}{89--94}.
\bibitem[{Nithin~Adidela and Sudhakar(????)}]{flowOpenFoam}
\bibinfo{author}{R.~S. Nithin~Adidela}, \bibinfo{author}{Y.~Sudhakar},
  \bibinfo{title}{Laminar flow over a circular cylinder simulation with
  openfoam v7},
  \bibinfo{howpublished}{\url{https://github.com/nithinadidela/circular-cylinder}},
  ????
\bibitem[{Driscoll et~al.(2014)Driscoll, Hale, and
  Trefethen}]{driscoll2014chebfun}
\bibinfo{author}{T.~A. Driscoll}, \bibinfo{author}{N.~Hale},
  \bibinfo{author}{L.~N. Trefethen}, \bibinfo{title}{Chebfun guide},
  \bibinfo{year}{2014}.
\bibitem[{Long et~al.(2018)Long, Lu, Ma, and Dong}]{long18a}
\bibinfo{author}{Z.~Long}, \bibinfo{author}{Y.~Lu}, \bibinfo{author}{X.~Ma},
  \bibinfo{author}{B.~Dong},
\newblock \bibinfo{title}{{PDE}-net: Learning {PDE}s from data},
\newblock in: \bibinfo{editor}{J.~Dy}, \bibinfo{editor}{A.~Krause} (Eds.),
  \bibinfo{booktitle}{Proceedings of the 35th International Conference on
  Machine Learning}, volume~\bibinfo{volume}{80} of
  \textit{\bibinfo{series}{Proceedings of Machine Learning Research}},
  \bibinfo{publisher}{PMLR}, \bibinfo{year}{2018}, pp.
  \bibinfo{pages}{3208--3216}. \URLprefix
  \url{https://proceedings.mlr.press/v80/long18a.html}.

\end{thebibliography}
\end{document}